\documentclass{article}

\usepackage{PRIMEarxiv}

\usepackage[utf8]{inputenc} 
\usepackage[T1]{fontenc}    
\usepackage{hyperref}       
\usepackage{url}            
\usepackage{booktabs}       
\usepackage{amsfonts}       
\usepackage{nicefrac}       
\usepackage{microtype}      
\usepackage{lipsum}
\usepackage{fancyhdr}       
\usepackage{graphicx}       
\graphicspath{{media/}}     

\usepackage{tikz}
\usepackage{IEEEtrantools}
\usepackage{amsmath}
\usepackage{amssymb}
\newcommand{\dd}{\mathop{}\!\mathrm{d}}
\usepackage{algorithm}
\usepackage{algorithmic}

\newtheorem{definition}{Definition}
\newtheorem{result}{Result}
\newtheorem{remark}{Remark}

\title{Differentially Private Transferrable Deep Learning with Membership-Mappings
\thanks{The research reported in this paper has been partly supported by EU Horizon 2020 Grant 826278 ``Securing Medical Data in Smart Patient-Centric Healthcare Systems'' (Serums), Austrian Research Promotion Agency (FFG) Grant 873979 ``Privacy Preserving Machine Learning for Industrial Applications'' (PRIMAL), and the Austrian Ministry for Transport, Innovation and Technology, the Federal Ministry for Digital and Economic Affairs, and the Province of Upper Austria in the frame of the COMET center SCCH.} 
}

\author{
  Mohit Kumar \\
  Software Competence Center Hagenberg, Austria 
  \and
  Faculty of Computer Science and Electrical Engineering, University of Rostock, Germany\\
  \texttt{\{mohit.kumar\}@uni-rostock.de} 
}

\begin{document}
\maketitle

\begin{abstract}
This paper considers the problem of differentially private semi-supervised transfer and multi-task learning. The notion of \emph{membership-mapping} has been developed using measure theory basis to learn data representation via a fuzzy membership function. An alternative conception of deep autoencoder, referred to as \emph{Conditionally Deep Membership-Mapping Autoencoder (CDMMA)}, is considered for transferrable deep learning. Under practice-oriented settings, an analytical solution for the learning of CDMMA can be derived by means of variational optimization. The paper proposes a transfer and multi-task learning approach that combines CDMMA with a tailored noise adding mechanism to achieve a given level of privacy-loss bound with the minimum perturbation of the data. Numerous experiments were carried out using MNIST, USPS, Office, and Caltech256 datasets to verify the competitive robust performance of the proposed methodology.
\end{abstract}


\section{Introduction}
The availability of high quality labelled data is crucial for the success of machine learning methods. While a single entity may not own massive amount of data, a collaboration among data-owners regarding sharing of knowledge extracted locally from their private data can be beneficial. The data privacy concerns and the legal requirements may not allow a centralization of the data from multiple sources. Thus, an interest in privacy-preserving machine learning with distributed training datasets arises. We consider the privacy-preserving distributed machine learning problem under a scenario that the knowledge extracted from a labelled training dataset (referred to as \emph{source domain}) is intended to improve the learning of a classifier trained using a dataset with both unlabelled and few labelled samples (referred to as \emph{target domain}) such that source and target domains are allowed to be heterogeneous. That is, source and target data samples are allowed to differ in their dimensions and no assumptions are made regarding statistical distributions of source and target data. The problem of \emph{privacy-preserving semi-supervised transfer learning} has previously been addressed in the literature from different prospects. We focus on the development of a method able to simultaneously deal with high-dimensional data and heterogeneous domains. 

\paragraph{State-of-the-art on Privacy-Preserving Transferrable Machine Learning} A lot of research has been carried out in the area of transfer learning. The heterogeneous data from source and target domain (i.e. source and target domains have different feature space and dimensions) can be transformed to a common subspace by using two different projection matrices. Existing supervised learning methods (e.g., SVM) can be then employed to learn the projection matrices and the target domain classifier~\cite{6587717}. It is possible to learn a transformation that maps feature points from one domain to another using cross-domain constraints formed by requiring that the transformation maps points from the same category (but different domain) near each other~\cite{Hoffman2014}. A study \cite{Herath_2017_CVPR} learns projections from each domain to a latent space via simultaneously minimizing a notion of domain variance while maximizing a measure of discriminatory power where Riemannian optimization techniques are used to match statistical properties between samples projected into the latent space from different domains. Another study \cite{7586038} proposes a regularized unsupervised optimal transportation model to perform the alignment of the representations in the source and target domains. The method in \cite{6247911} uses geodesic flow to construct an infinite-dimensional feature space that assembles information on the source domain, on the target domain, and on \emph{phantom} domains interpolating between source and target domains. Inner products in infinite-dimensional feature space give rise to a kernel function facilitating the construction of any kernelized classifiers. Another approach is of an adaptation of source model to the target domain via iteratively deleting source-domain samples and adapting the model gradually to the target-domain instances \cite{10.1109/TPAMI.2009.57}. Boosting-based learning algorithms can be also used to adaptively assign the training weights to source and target samples based on their relevance in the training of the classifier \cite{10.1145/1273496.1273521}. Bayesian learning can be a framework to study transfer learning through modeling of a joint prior probability density function for feature-label distributions of the source and target domains~\cite{8362683}. Deep learning framework is another promising research direction explored for transfer learning~\cite{pmlr-v37-long15,10.5555/3157096.3157112,10.5555/2946645.2946704}.   

The datasets may contain sensitive information that need to be protected from \emph{model inversion} attack~\cite{Fredrikson:2015:MIA:2810103.2813677} and from adversaries with an access to model parameters and knowledge of the training procedure. This goal has been addressed within the framework of differential privacy~\cite{10.1145/2976749.2978318,Phan:2016:DPP:3015812.3016005}. Differential Privacy~\cite{10.1007/11761679_29,DBLP:journals/fttcs/DworkR14} is a formalism to quantify the degree to which the privacy for each individual in the dataset is preserved while releasing the output of a data analysis algorithm. Differential privacy provides a guarantee that an adversary, by virtue of presence or absence of an individual's data in the dataset, would not be able to draw any conclusions about an individual from the released output of the analysis algorithm. This guarantee is achieved by means of a randomization of the data analysis process. In the context of machine learning, randomization is carried out via either adding random noise to the input or output of the machine learning algorithm or modifying the machine learning algorithm itself. A limited number of studies exist on differentially private semi-supervised transfer learning. The authors in \cite{10.1007/s10994-013-5396-x} suggest an importance weighting mechanism to preserve the differential privacy of a private dataset via computing and releasing a weight for each record in an existing public dataset such that computations on public dataset with weights is approximately equivalent to computations on private dataset. The importance weighting mechanism is adapted in~ \cite{DBLP:conf/pkdd/WangGB18} to determine the weight of a source hypothesis in the process of constructing informative Bayesian prior for logistic regression based target model. \cite{conf/iclr/PapernotAEGT17} introduces \emph{private aggregation of teacher ensembles} approach where an ensemble of \emph{teacher} models is trained on disjoint subsets of the sensitive data and a \emph{student} model learns to predict an output chosen by noisy voting among all of the teachers. Another approach \cite{8215544,Xie2018DifferentiallyPG,10.1145/3134428} is to construct a differentially private unsupervised generative model for generating a synthetic version of the private data, and then releases the synthetic data for a non-private learning. This technique is capable of effectively handling high-dimensional data in differential privacy setting. The study in \cite{10.1093/bioinformatics/btz373} uses a large public dataset to learn a dimension-reducing representation mapping which is then applied on private data to obtain a low-dimensional representation of the private data followed by the learning of a differentially private predictor. However, these methods don't consider the heterogeneous domains.    

Differential privacy preserves the privacy of the training dataset via adding random noise to ensure that an adversary can not infer any single data instance by observing model parameters or model outputs. We follow the \emph{input perturbation} method where noise is added to original data to achieve $(\epsilon,\delta)-$differential privacy of any subsequent computational algorithm processing the perturbed data. However, the injection of noise into data would in general result in a loss of algorithm's accuracy. Therefore, design of a noise injection mechanism achieving a good trade-off between privacy and accuracy is a topic of interest~\cite{DBLP:journals/corr/abs-1809-10224,DBLP:journals/corr/abs-1805-06530,doi:10.1137/09076828X,Gupte:2010:UOP:1807085.1807105,7345591,7093132,7353177}. The authors in~\cite{Kumar/IWCFS2019} derive the probability density function of noise that minimizes the expected noise magnitude together with satisfying the sufficient conditions for $(\epsilon,\delta)-$differential privacy. This noise adding mechanism was applied for differentially private distributed deep learning in~\cite{10.1145/3386392.3399562,KUMAR202187}. In this study, the optimal noise adding mechanism of~\cite{Kumar/IWCFS2019} is applied for differentially private semi-supervised transfer learning.

Deep neural networks outperform classical machine learning techniques in a wide range of applications but their training requires a large amount of data. The issues, such as determining the optimal model structure, smaller training dataset, and iterative time-consuming nature of numerical learning algorithms, are inherent to the neural networks based parametric deep models. The nonparametric approach on the other hand can be promising to address the issue of optimal choice of model structure. However, an analytical solution instead of iterative gradient-based numerical algorithms will be still desired for the learning of deep models. These motivations have led to the development of a nonparametric deep model~\cite{8888203,9216097,KUMAR20211} that is learned analytically for representing data points. The study in~\cite{8888203,9216097} introduces the concept of \emph{Student-t fuzzy-mapping} which is about representing mappings through a fuzzy set with Student-t type membership function such that the dimension of membership function increases with an increasing data size. A relevant result is that a deep autoencoder model formed via a composition of finite number of nonparametric fuzzy-mappings can be learned analytically. However, \cite{8888203,9216097} didn't provide a formal mathematical framework for the conceptualization of so-called fuzzy-mapping. The study in~\cite{10.1007/978-3-030-87101-7_13} provides to fuzzy-mapping a \emph{measure-theoretic} conceptualization and refers it to as \emph{membership-mapping}. The interpolation property of the membership-mapping was exploited for data representation learning for developing an analytical approach to the variational learning of a membership-mappings based data representation model. Further,~\cite{10.1007/978-3-030-87101-7_14} uses membership-mapping as the building block of deep models. An alternative idea of deep autoencoder, referred to as Bregman Divergence Based Conditionally Deep Autoencoder (that consists of layers such that each layer learns data representation at certain abstraction level through a membership-mappings based autoencoder), was presented in~\cite{10.1007/978-3-030-87101-7_14}. In this study, we leverage the data representation learning capability of the conditionally deep autoencoder for privacy-preserving semi-supervised transferrable deep learning. The conditionally deep autoencoder considered in this study is referred to as conditionally deep membership-mapping autoencoder to emphasize membership-mapping serving as the building block of the deep model.

\paragraph{Motivation} The motivation of this study is derived from the ambition of developing a differentially private semi-supervised transfer and multi-task learning framework that
\begin{enumerate}
\item[R1:] is capable of handling high-dimensional data and heterogeneity of domains;
\item[R2:] optimizes the differential private noise adding mechanism such that for a given level of privacy, the perturbation in the data is as small as possible; 
\item[R3:] allows learning of the target domain model without requiring an access to source domain private training data;
\item[R4:] ensures that a high level of privacy (i.e. sufficiently low value of privacy-loss bound) would not degrade the learning performance;  
\item[R5:] allows employing deep models in source and target domains so that data features at different abstraction levels can be used to transfer knowledge across domains;
\item[R6:] follows the analytical approach~\cite{8888203,9216097,KUMAR20211,10.1007/978-3-030-87101-7_13,10.1007/978-3-030-87101-7_14} to the learning of deep models while addressing the issues related to optimal choice of model structure.
\end{enumerate}
To the best knowledge of authors, there does not exist any study in the literature addressing sufficiently simultaneously all of the aforementioned six requirements (i.e. R1-R6). We present in this study a novel approach to differentially private transfer and multi-task learning that fulfills all of the requirements.

\paragraph{Proposed Method}
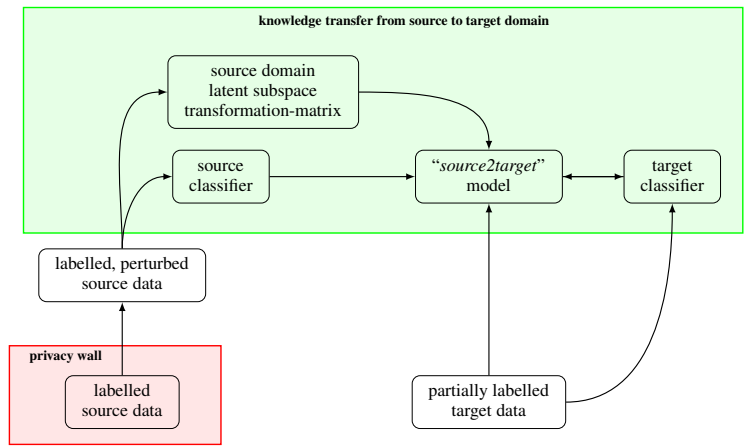
\begin{figure}[bt]
\centering
\scalebox{0.75}{
\begin{tikzpicture}[scale=1]
   \path[fill=green!10](-1.25,2.5)--(11.5,2.5)--(11.5,6.5)--(-1.25,6.5)--cycle;
   \draw[green,line width=0.25mm](-1.25,2.5)--(11.5,2.5)--(11.5,6.5)--(-1.25,6.5)--cycle;
    \draw (5.5,6.25) node[]{\bfseries \scriptsize $\begin{array}{c} \mbox{knowledge transfer from source to target domain} \end{array}$};
\draw (0.5,1.75) node[rounded corners,draw](n1){\footnotesize $\begin{array}{c} \mbox{labelled, perturbed} \\ \mbox{source data} \end{array}$};
 \path[red, fill=red!10](-1.5,-1.25)--(2.25,-1.25)--(2.25,0.5)--(-1.5,0.5)--cycle;
  \draw[red,line width = 0.25mm] (-1.5,-1.25)--(2.25,-1.25)--(2.25,0.5)--(-1.5,0.5)--cycle;
 \draw (-0.5,0.3) node[]{\bfseries \scriptsize privacy wall};
 \draw (2.25,3.5) node[draw, rounded corners](n2){\footnotesize $\begin{array}{c} \mbox{source} \\ \mbox{classifier} \end{array}$};
 \draw[-latex,line width=0.2mm] (n1) to [out=90,in=180] (n2); 
 \draw (7,3.5) node[draw, rounded corners](n4){\footnotesize $\begin{array}{c} \mbox{``\emph{source2target}''} \\ \mbox{model} \end{array}$}; 
 \draw (7,-0.5) node[rounded corners,draw](n5){\footnotesize $\begin{array}{c} \mbox{partially labelled} \\ \mbox{target data} \end{array}$};
  \draw[-latex,line width=0.2mm] (n2) to [out=0,in=180] (n4);   
    \draw[-latex,line width=0.2mm] (n5) to [out=90,in=-90] (n4);  
  \draw (10.25,3.5) node[draw, rounded corners](n6){\footnotesize $\begin{array}{c} \mbox{target} \\ \mbox{classifier}  \end{array}$}; 
     \draw[-latex,line width=0.2mm] (n5) to [out=0,in=-90] (n6); 
  \draw[-latex,line width=0.2mm] (n4) to [out=0,in=180] (n6);    
\draw[-latex,line width=0.2mm] (n6) to [out=180,in=0] (n4);  
  \draw (3,5) node[draw, rounded corners](n7){\footnotesize $\begin{array}{c} \mbox{source domain} \\ \mbox{latent subspace} \\ \mbox{transformation-matrix}  \end{array}$};
 \draw[-latex,line width=0.2mm] (n1) to [out=90,in=180] (n7);  
  \draw[-latex,line width=0.2mm] (n7) to [out=0,in=90] (n4);  
 \draw (0.5,-0.5) node[rounded corners,draw](n8){\footnotesize $\begin{array}{c} \mbox{labelled} \\ \mbox{source data} \end{array}$};   
\draw[-latex,line width=0.2mm] (n8) to [out=90,in=-90] (n1);  
\end{tikzpicture}
}
\caption{The proposed approach to privacy-preserving semi-supervised transfer and multi-task learning.}
\label{fig_basic_idea}
\end{figure} 
The basic idea of the proposed approach is stated as in Fig.~\ref{fig_basic_idea}. The method is as follows:   
\begin{itemize}
\item An optimal differentially private noise adding mechanism is used to perturb the source dataset for preserving its privacy. The perturbed source data is used for the learning of classifier and for the computation of other parameters required for transferring knowledge from source to target domain. 
\item The classifiers consist of \emph{Conditionally Deep Membership-Mapping Autoencoder (CDMMA)} based compositions. A multi-class classifier is presented that employs a parallel composition of CDMMAs to learn data representation for each class. An analytical approach is presented for the learning of the CDMMA. 
\item Since differential privacy will remain immune to any post-processing of noise added data samples, the perturbed source dataset is used to
\begin{itemize}
\item build a differentially private source domain classifier,
\item compute a differentially private source domain latent subspace transformation-matrix.
\end{itemize}
\item The target domain classifier is learned adaptively in a manner that higher-level data features are used during initial iterations for updating the classifier parameters and as the number of iterations increases more and more lower-level data features are intended to be included in the process of updating the classifier parameters.
\item The knowledge from source to target domain is transferred via
\begin{enumerate}
\item  building a ``\emph{source2target}'' model that uses variational membership-mappings to define a transformation from source domain data space to the target domain data space,
\item combining both source  and target domain classifiers with source2target model for a transfer and \emph{multi-task} learning scenario.          
\end{enumerate} 
\item Since no flow of data/information occurs from target to source domain, no privacy-preserving mechanism is used to protect the target data.  
\end{itemize}
\paragraph{Novelty} The source and target models in our methodology employ CDMMAs. The classical deep autoencoder consists of two symmetrical networks of multiple layers such that first network represents the encoding and second network represents the decoding. However, the CDMMA considered in this study is composed of layers such that each layer learns data representation at certain abstraction level through a \emph{membership-mapping autoencoder}~\cite{10.1007/978-3-030-87101-7_14}. Motivated by fuzzy theory,  the notion of membership-mapping has been introduced~\cite{10.1007/978-3-030-87101-7_13} for representing data points through attribute values. Following~\cite{10.1007/978-3-030-87101-7_14}, we consider the approach of using membership-mapping as the building block of deep models. The motivation behind this approach is derived from the facts that an analytical learning solution could be derived for membership-mappings via variational optimization methodology and thus the typical issues associated to parametric deep models (such as determining the optimal model structure, smaller training dataset, and iterative time-consuming nature of numerical learning algorithms) will be automatically addressed. The novelty of this study is that an analytical approach to the learning of CDMMA, facilitating high-dimensional data representation learning at varying abstraction levels across CDMMA's different layers, will be applied for transferring knowledge from source to target domain in a privacy-preserving manner. 
\paragraph{Contributions} Leveraging the data representation capabilities of membership-mappings is central to our approach, and thus development of an efficient algorithm for their learning is required. Although~\cite{10.1007/978-3-030-87101-7_13} provides an algorithm for the variational learning of membership-mappings via following the approach of~\cite{8888203,9216097,KUMAR20211}, there remain the following limitations:
\begin{enumerate}
\item The derivation of the learning solution via following the approach of~\cite{8888203,9216097,KUMAR20211} is complex and difficult to follow.
\item The suggested learning algorithm involves several free parameters.
\end{enumerate}
In this study, we address these two limitations via deriving a more simple and elegant solution for the variational learning of membership-mappings with an additional advantage of automatic determination of free parameters related to the learning algorithm. Thus, one of the contributions of this study is to present an efficient algorithm for the variational learning of membership-mappings. More importantly, the major contribution is to address all of the posed requirements of differentially private transferrable deep learning: R1-R6. Sufficient experimentation has been provided on benchmark problems to demonstrate the merit of the proposed approach. 
\paragraph{Organization} Section~\ref{sec_2} provides the necessary background on membership-mappings. An analytical solution to the variational learning of CDMMAs is presented in section~\ref{sec_3}. The analytical solution derived in section~\ref{sec_3} facilitates in section~\ref{sec_4} a differentially private semi-supervised transfer and multi-task learning method implementing an optimal noise adding mechanism. The experiments in section~\ref{sec_5} provide demonstrative examples (on MNIST and USPS datasets) and the comparison with standard techniques (on Office and Caltech256 datasets). Finally, concluding remarks are given in section~\ref{sec_conclusion}.    
\section{Background}\label{sec_2}
\subsection{Notations}
\begin{itemize}
\item Let $n,N,p,M \in \mathbb{N}$. 
\item Let $\mathcal{B}(\mathbb{R}^N)$ denote the \emph{Borel $\sigma-$algebra} on $\mathbb{R}^N$, and let $\lambda^N$ denote the \emph{Lebesgue measure} on $\mathcal{B}(\mathbb{R}^N)$. 
\item Let $(\mathcal{X}, \mathcal{A} , \rho)$ be a probability space with unknown probability measure $\rho$. 
\item Let us denote by $\mathcal{S}$ the set of finite samples of data points drawn i.i.d. from $\rho$, i.e.,
\begin{IEEEeqnarray}{rCl}  
\mathcal{S} & := &  \{ (x^i  \sim \rho )_{i=1}^N \; | \; N \in \mathbb{N} \}.
\end{IEEEeqnarray} 
\item For a sequence $\mathrm{x} = (x^1,\cdots,x^N) \in \mathcal{S}$, let $|\mathrm{x}|$ denote the cardinality i.e. $|\mathrm{x}| = N$.     
\item If $\mathrm{x} = (x^1, \cdots, x^N),\; \mathrm{a}= (a^1, \cdots, a^M) \in \mathcal{S}$, then $\mathrm{x} \wedge \mathrm{a}$ denotes the concatenation of the sequences $\mathrm{x}$ and $\mathrm{a}$, i.e., $\mathrm{x} \wedge \mathrm{a} = (x^1, \ldots, x^N, a^1, \ldots, a^M)$.
\item Let us denote by $\mathbb{F}(\mathcal{X})$ the set of $\mathcal{A}$-$\mathcal{B}(\mathbb{R})$ measurable functions $f:\mathcal{X} \rightarrow \mathbb{R}$, i.e.,
\begin{IEEEeqnarray}{rCl}  
\mathbb{F}(\mathcal{X}) & := &  \{ f:\mathcal{X} \rightarrow \mathbb{R}  \; | \;  \mbox{$f$ is $\mathcal{A}$-$\mathcal{B}(\mathbb{R})$ measurable}\}.
\end{IEEEeqnarray} 
\item For convenience, the values of a function $f \in \mathbb{F}(\mathcal{X})$ at points in the collection $\mathrm{x} = (x^1,\cdots,x^N)$ are represented as $f(\mathrm{x})=(f(x^1),\cdots,f(x^N))$.
\item For a given $\mathrm{x} \in \mathcal{S}$ and $A\in \mathcal{B}(\mathbb{R}^{|\mathrm{x}|})$, the cylinder set $\mathcal{T}_{\mathrm{x}}(A)$ in $\mathbb{F}(\mathcal{X})$ is defined as 
\begin{IEEEeqnarray}{rCl}
\mathcal{T}_{\mathrm{x}}(A) &: = & \{ f \in \mathbb{F}(\mathcal{X}) \; | \; f(\mathrm{x}) \in A   \}.
\end{IEEEeqnarray}
\item Let $\mathcal{T}$ be the family of cylinder sets defined as
\begin{IEEEeqnarray}{rCl}
\mathcal{T} & := & \left\{ \mathcal{T}_{\mathrm{x}}(A)\; | \; A \in \mathcal{B}(\mathbb{R}^{|\mathrm{x}|}),\; \mathrm{x} \in \mathcal{S} \right \}.
\end{IEEEeqnarray} 
\item Let $\sigma(\mathcal{T})$ be the $\sigma$-algebra generated by $\mathcal{T}$.
\item Given two $\mathcal{B}(\mathbb{R}^N)-\mathcal{B}(\mathbb{R})$ measurable mappings, $g:\mathbb{R}^N \rightarrow \mathbb{R}$ and $\mu:\mathbb{R}^N \rightarrow \mathbb{R}$, the weighted average of $g(\mathrm{y})$ over all $\mathrm{y} \in \mathbb{R}^{N}$, with $\mu(\mathrm{y})$ as the weighting function, is computed as   
\begin{IEEEeqnarray}{rCl}
\label{eq_738118.427179} \left< g \right>_{\mu}& := & \frac{1}{ \int_{\mathbb{R}^{N}}  \mu(\mathrm{y})\, \dd\lambda^{N}(\mathrm{y})} \int_{\mathbb{R}^{N}} g(\mathrm{y}) \mu(\mathrm{y})\, \dd \lambda^{N}(\mathrm{y}).
\end{IEEEeqnarray} 
\end{itemize}
\subsection{A Review of Measure Theoretic Conceptualization of Membership-Mappings}
\subsubsection{Representation of Samples via Attribute Values}
Consider a given observation $x \in \mathcal{X}$, a data point $\tilde{x} \in \mathcal{X}$, and a mapping $\mathbf{A}_{x,f}(\tilde x) = (\zeta_{x} \circ f) (\tilde x)$ composed of two mappings $f: \mathcal{X} \rightarrow \mathbb{R}$ and $\zeta_x:  \mathbb{R} \rightarrow [0,1]$. $f \in \mathbb{F}(\mathcal{X})$ can be interpreted as physical measurement (e.g., temperature), and $\zeta_x (f(\tilde x))$ as degree to which $\tilde x$ matches the attribute under consideration, e.g. ``\emph{hot}'' where e.g. $x$ is a representative sample of ``\emph{hot}''. This concept is extended to sequences of data points in order to evaluate how much a sequence $\tilde{\mathrm{x}} = (\tilde x^1, \ldots, \tilde x^N) \in \mathcal{S}$ matches to the attribute induced by observed sequence $\mathrm{x} = (x^1, \ldots, x^N) \in \mathcal{S}$ w.r.t. the feature $f$ via defining
\begin{IEEEeqnarray}{rCl}
\mathbf{A}_{\mathrm{x},f}(\tilde{\mathrm{x}}) & = & (\zeta_{\mathrm{x}} \circ f)(\tilde{\mathrm{x}}) \\
& = & \zeta_{\mathrm{x}} (f(\tilde x^1), \ldots, f(\tilde x^N)),
\end{IEEEeqnarray} 
where the membership functions $\zeta_{\mathrm{x}}:\mathbb{R}^{|\mathrm{x}|} \rightarrow [0,1]$, $\mathrm{x} \in \mathcal{S}$, satisfy the following properties:
\begin{description}
    \item[Nowhere Vanishing:] $\zeta_{\mathrm{x}}(\mathrm{y}) > 0$ for all $\mathrm{y} \in \mathbb{R}^{|\mathrm{x}|}$, i.e.,
\begin{IEEEeqnarray}{rCl}
    \label{eq:supp}
     \mbox{supp}[\zeta_{\mathrm{x}}] & = & \mathbb{R}^{|\mathrm{x}|}.
\end{IEEEeqnarray}   
    \item[Positive and Bounded Integrals:] the functions $\zeta_{\mathrm{x}}$ are absolutely continuous and Lebesgue integrable over the whole domain such that for all $\mathrm{x}\in \mathcal{S}$ we have
     \begin{eqnarray}
    \label{eq:positive}
   0 < \int_{\mathbb{R}^{|\mathrm{x}|}} \zeta_{\mathrm{x}}\, \dd \lambda^{|\mathrm{x}|}  < \infty.
   \end{eqnarray}
   \item[Consistency of Induced Probability Measure:] the membership function induced probability measures $\mathbb{P}_{\zeta_{\mathrm{x}}}$, defined on any $A \in \mathcal{B}(\mathbb{R}^{|\mathrm{x}|})$, as
\begin{IEEEeqnarray}{rCl}
\mathbb{P}_{\zeta_{\mathrm{x}}}(A) & := &  \frac{1}{ \int_{\mathbb{R}^{|\mathrm{x}|}} \zeta_{\mathrm{x}}\, \dd \lambda^{|\mathrm{x}|}} \int_{A} \zeta_{\mathrm{x}}\, \dd\lambda^{|\mathrm{x}|}
\end{IEEEeqnarray}  
are consistent in the sense that for all $\mathrm{x},\;\mathrm{a} \in \mathcal{S}$:
\begin{IEEEeqnarray}{rCl}
\label{eq_738083.390026} \mathbb{P}_{\zeta_{\mathrm{x} \wedge \mathrm{a}}}(A \times \mathbb{R}^{|\mathrm{a}|}) & = & \mathbb{P}_{\zeta_{\mathrm{x}}}(A). 
\end{IEEEeqnarray}  
\end{description}
The collection of membership functions satisfying aforementioned assumptions is denoted by 
\begin{IEEEeqnarray}{rCl}
\Theta & := & \{ \zeta_{\mathrm{x}}:\mathbb{R}^{|\mathrm{x}|} \rightarrow [0,1] \; | \; (\ref{eq:supp}),  (\ref{eq:positive}), (\ref{eq_738083.390026}),\; \mathrm{x} \in \mathcal{S}\}.
\end{IEEEeqnarray}
\subsubsection{Measure Space}
It is shown in~\cite{10.1007/978-3-030-87101-7_13} that $(\mathbb{F}(\mathcal{X}),\sigma(\mathcal{T}),\mathbf{p})$ is a measure space and the probability measure $\mathbf{p}$ is defined as       
 \begin{IEEEeqnarray}{rCl}
\label{eq_738082.742718}\mathbf{p}(\mathcal{T}_{\mathrm{x}}(A) ) & := &  \mathbb{P}_{\zeta_{\mathrm{x}}}(A)
\end{IEEEeqnarray} 
where $\zeta_{\mathrm{x}} \in \Theta$, $\mathrm{x} \in \mathcal{S}$, $A\in \mathcal{B}(\mathbb{R}^{|\mathrm{x}|})$, and $\mathcal{T}_{\mathrm{x}}(A) \in \mathcal{T}$. It follows from~\cite{10.1007/978-3-030-87101-7_13} that for a given $\mathcal{B}(\mathbb{R}^{|\mathrm{x}|})-\mathcal{B}(\mathbb{R})$ measurable mapping $g:\mathbb{R}^{|\mathrm{x}|} \rightarrow \mathbb{R}$, expectation of $(g\circ f)(\mathrm{x})$ over $f \in \mathbb{F}(\mathcal{X})$ w.r.t. probability measure $\mathbf{p}$ is given as 
\begin{IEEEeqnarray}{rCl} 
\label{eq_738116.460670} \mathbb{E}_{\mathbf{p}}[(g\circ \cdot)(\mathrm{x})]  & = & \left< g \right >_{\zeta_{\mathrm{x}}}.
\end{IEEEeqnarray} 
The significance of equality~(\ref{eq_738116.460670}) is to allow calculating averages over all real valued functions belonging to $\mathbb{F}(\mathcal{X})$ via simply computing a weighted average. 
\subsubsection{Student-t Membership-Mapping}
\begin{definition}[Student-t Membership-Mapping]\label{def_student_t_set_membership_mapping}
A Student-t membership-mapping, $\mathcal{F} \in \mathbb{F}(\mathcal{X})$, is a mapping with input space $\mathcal{X} = \mathbb{R}^n$ and a membership function $\zeta_{\mathrm{x}} \in \Theta$ that is Student-t like:
\begin{IEEEeqnarray}{rCl}
\label{eq_student_t_membership} 
\label{eq_738098.751419}\zeta_{\mathrm{x}}(\mathrm{y}) & = & \left(1 + 1/(\nu - 2) \left( \mathrm{y} - \mathrm{m}_{\mathrm{y}} \right)^T K^{-1}_{\mathrm{x}\mathrm{x}} \left( \mathrm{y}- \mathrm{m}_{\mathrm{y}}\right) \right)^{-\frac{\nu+|\mathrm{x}|}{2}}
\end{IEEEeqnarray} 
where $\mathrm{x} \in \mathcal{S}$, $\mathrm{y} \in \mathbb{R}^{|\mathrm{x}|}$, $\nu \in \mathbb{R}_{+}\setminus [0,2]$ is the degrees of freedom, $\mathrm{m}_{\mathrm{y}} \in \mathbb{R}^{|\mathrm{x}|}$ is the mean vector, and $K_{\mathrm{x}\mathrm{x}} \in \mathbb{R}^{|\mathrm{x}| \times |\mathrm{x}|}$ is the covariance matrix with its $(i,j)-$th element given as 
\begin{IEEEeqnarray}{rCl}
\label{738026.844153}  (K_{\mathrm{x}\mathrm{x}})_{i,j} & = & kr(x^i,x^j) 
 \end{IEEEeqnarray}  
where $kr: \mathbb{R}^n \times \mathbb{R}^n \rightarrow \mathbb{R}$ is a positive definite kernel function defined as 
\begin{IEEEeqnarray}{rCl}
\label{eq_membership1003_3} kr(x^{i},x^{j}) & = &  \sigma^2 \exp \left(-0.5\sum_{k = 1}^{n} w_{k} \left |  x^{i}_k - x^{j}_k \right |^2\right)
 \end{IEEEeqnarray}  
where $x_k^i$ is the $k-$th element of $x^i$, $\sigma^2$ is the variance parameter, and $w_{k} \geq 0$ (for $k \in \{1,\cdots,n\}$).
\end{definition}
It is shown in~\cite{10.1007/978-3-030-87101-7_13} that membership function as defined in (\ref{eq_738098.751419}) satisfies the consistency condition~(\ref{eq_738083.390026}).  
\subsubsection{Interpolation}
For a zero-mean Student-t membership-mapping $\mathcal{F} \in \mathbb{F}(\mathbb{R}^n)$, let $\mathrm{x} = \{x^i \in \mathbb{R}^n \; | \; i \in \{1,\cdots,N\}\}$ be a given set of input points and the corresponding mapping outputs are represented by the vector $\mathrm{f}  :=  (\mathcal{F}(x^1), \cdots, \mathcal{F}(x^N))$. Let $\mathrm{a} = \{a^m\;|\; a^m \in \mathbb{R}^n,\; m \in \{1,\cdots,M \}  \}$ be the set of auxiliary inducing points and the mapping outputs corresponding to auxiliary inducing inputs are represented by the vector $\mathrm{u}  :=  (\mathcal{F}(a^1) , \cdots , \mathcal{F}(a^M))$. It follows from~\cite{10.1007/978-3-030-87101-7_13} that $\mathrm{f}$, based upon the interpolation on elements of $\mathrm{u}$, could be represented by means of a membership function, $\mu_{\mathrm{f};\mathrm{u}}:\mathbb{R}^N \rightarrow [0,1]$, defined as 
 \begin{IEEEeqnarray}{rCl}
\mu_{\mathrm{f};\mathrm{u}}(\tilde{\mathrm{f}})  & := & \left( 1 + \frac{1}{\nu + M - 2} (\tilde{\mathrm{f}} -  \bar{m}_{\mathrm{f}})^T    \left( \frac{\nu + (\mathrm{u})^T (K_{\mathrm{a}\mathrm{a}})^{-1} \mathrm{u}  - 2}{\nu + M - 2} \bar{K}_{\mathrm{x}\mathrm{x}} \right)^{-1}(\tilde{\mathrm{f}} -  \bar{m}_{\mathrm{f}}) \right)^{-\frac{\nu+M+N}{2}} \\
\label{eq_738124.571185} \bar{m}_{\mathrm{f}} & = &  K_{\mathrm{x}\mathrm{a}} (K_{\mathrm{a}\mathrm{a}})^{-1}  \mathrm{u} \\     
\bar{K}_{\mathrm{x}\mathrm{x}} & = & K_{\mathrm{x}\mathrm{x}} - K_{\mathrm{x}\mathrm{a}} (K_{\mathrm{a}\mathrm{a}})^{-1} K_{\mathrm{x}\mathrm{a}}^T,
\end{IEEEeqnarray}    
where $K_{\mathrm{a}\mathrm{a}} \in \mathbb{R}^{M \times M}$ and $K_{\mathrm{x}\mathrm{a}}  \in \mathbb{R}^{N \times M}$ are matrices with their $(i,j)-$th elements given as
\begin{IEEEeqnarray}{rCl}
\label{eq_membership1004_2} \left( K_{\mathrm{a}\mathrm{a}} \right)_{i,j} & = & kr(a^{i},a^{j}) \\
\label{eq_738497.4922} \left( K_{\mathrm{x}\mathrm{a}} \right)_{i,j} & = & kr(x^{i},a^{j})
 \end{IEEEeqnarray} 
where $kr: \mathbb{R}^n \times \mathbb{R}^n \rightarrow \mathbb{R}$ is a positive definite kernel function defined as in (\ref{eq_membership1003_3}).

Here, the pair $(\mathbb{R}^N,\mu_{\mathrm{f};\mathrm{u}})$ constitutes a fuzzy set and $\mu_{\mathrm{f};\mathrm{u}}(\tilde{\mathrm{f}})$ is interpreted as the degree to which $\tilde{\mathrm{f}}$ matches an attribute induced by $\mathrm{f}$ for a given $\mathrm{u}$.     

\subsection{A Review of Conditionally Deep Membership-Mapping Autoencoders}\label{subsec_reviw_CDMMA}
\begin{figure}
\centering
\scalebox{0.7}{
\begin{tikzpicture}
 \draw[fill=gray!5] (0.5,-4.5) rectangle (12.5,1); 
  \draw[arrows=->](0,0)--(1.5,0) node[below,pos=0]{$y$};
   \draw (1.5,-0.5) rectangle (2.5,0.5) node[pos=0.5]{$\mathcal{G}_1(\cdot)$};  
     \draw[arrows=->](2.5,0)--(3.5,0) node[above, pos = 0.5]{$y^1$};
 \draw (3.5,-0.5) rectangle (4.5,0.5) node[pos=0.5]{$\mathcal{G}_2(\cdot)$};  
      \draw[arrows=->](4.5,0)--(5.5,0) node[above, pos = 0.5]{$y^2$};
 \draw[dashed](5.5,0)--(7.5,0);
       \draw[arrows=->](7.5,0)--(8.5,0);
 \draw (8.5,-0.5) rectangle (9.5,0.5) node[pos=0.5]{$\mathcal{G}_L(\cdot)$};  
 \draw[arrows=->](9.5,0)--(10.5,0) node[above, pos = 0.5]{$y^L$};  
 \draw (10.5,-4.1) rectangle (12,0.5) node[pos=0.5]{$\begin{array}{c}\mbox{output} \\ \mbox{layer} \end{array}$};      
 \draw[](5,0)--(5,-2.25);
 \draw[dashed](10,-0.25)--(10,-1.5);
  \draw[arrows=->](5,-2.25)--(10.5,-2.25) node[above, pos = 0.9]{$y^{2}$}; 
   \draw[](3,0)--(3,-3);
     \draw[arrows=->](3,-3)--(10.5,-3) node[above, pos = 0.925]{$y^{1}$}; 
 \draw[](1,0)--(1,-3.75);    
 \draw[arrows=->](1,-3.75)--(10.5,-3.75) node[above, pos = 0.935]{$y$}; 
   \draw[arrows=->](12,-1.8)--(13,-1.8) node[below, pos = 1]{$y^{l^*}$}; 
 \draw (5.5,-7.5) rectangle (7.5,-6.5) node[pos=0.5]{$\mathcal{D}(\cdot)$};  
    \draw[arrows=->](4,-7)--(5.5,-7) node[left,pos=0]{$y$};
  \draw[arrows=->](7.5,-7)--(8.5,-7) node[right, pos = 1]{$y^{l^*}$}; 
   \path[fill=gray!10](0.5,-4.5)--(5.5,-6.5)--(7.5,-6.5)--(12.5,-4.5)--cycle;
\end{tikzpicture}
}
\caption{The structure of an $L-$layered conditionally deep autoencoder consisting of a nested compositions of membership-mapping autoencoders.}
\label{fig_deep_autoencoder}
\end{figure}
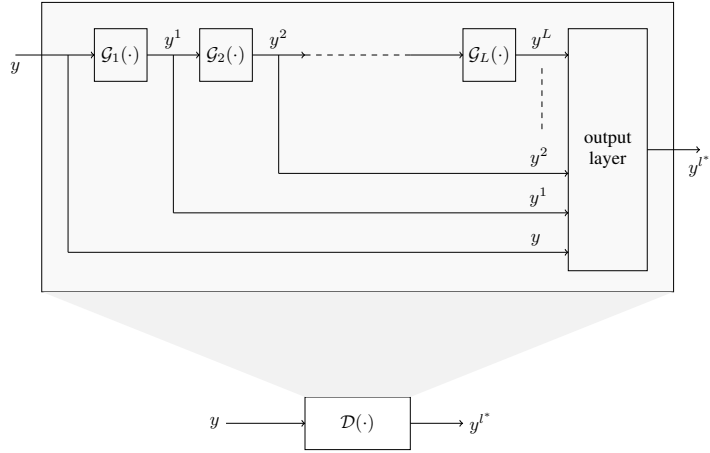

Bregman divergence based conditionally deep autoencoders were introduced in~\cite{10.1007/978-3-030-87101-7_14}. In this study, we consider a special case of Bregman divergence corresponding to the squared Euclidean norm for the conditionally deep autoencoders.
\begin{definition}[Membership-Mapping Autoencoder~\cite{10.1007/978-3-030-87101-7_14}]\label{def_SFMA}
A membership-mapping autoencoder, $\mathcal{G}:\mathbb{R}^p \rightarrow \mathbb{R}^p$, maps an input vector $y \in \mathbb{R}^p$ to $\mathcal{G}(y) \in \mathbb{R}^p$ such that 
 \begin{IEEEeqnarray}{rCl}
\label{eq_added_before_publication_1}  \mathcal{G}(y) &    := &  \left[\begin{IEEEeqnarraybox*}[][c]{,c/c/c,}  \mathcal{F}_1(Py) & \cdots &  \mathcal{F}_p(Py)
 \end{IEEEeqnarraybox*} \right]^T, 
\end{IEEEeqnarray} 
where $\mathcal{F}_j$ ($j \in \{1,2,\cdots,p\}$) is a Student-t membership-mapping (Definition~\ref{def_student_t_set_membership_mapping}), $P \in \mathbb{R}^{n \times p} (n \leq p)$ is a matrix such that the product $Py$ is a lower-dimensional encoding for $y$. That is, membership-mapping autoencoder first projects the input vector onto a lower dimensional subspace and then constructs the output vector through Student-t membership-mappings. 
\end{definition}
\begin{definition}[Conditionally Deep Membership-Mapping Autoencoder (CDMMA)~\cite{10.1007/978-3-030-87101-7_14}]\label{def_deep_autoencoder}
A conditionally deep membership-mapping autoencoder, $\mathcal{D}:\mathbb{R}^p \rightarrow \mathbb{R}^p$, maps a vector $y \in \mathbb{R}^p$ to $\mathcal{D}(y) \in \mathbb{R}^p$ through a nested composition of finite number of membership-mapping autoencoders such that
 \begin{IEEEeqnarray}{rCl}
 y^l & = & (\mathcal{G}_l \circ \cdots \circ \mathcal{G}_2\circ \mathcal{G}_1)(y), \; \forall l \in \{1,2,\cdots,L \}\\
 l^* & = & \arg\;\min_{l \: {\in} \: \{1,2,\cdots,L \}}\; \| y -  y^l \|^2 \\
  \mathcal{D}(y) & = & y^{l^*},
\end{IEEEeqnarray}  
where $\mathcal{G}_l(\cdot)$ is a membership-mapping autoencoder (Definition~\ref{def_SFMA}); $y^l$ is the output of $l-$th layer representing input vector $y$ at certain abstraction level such that $y^1$ is least abstract representation and $y^L$ is most abstract representation of the input vector; and the autoencoder output $\mathcal{D}(y)$ is equal to the output of the layer re-constructing the given input vector as good as possible where re-construction error is measured in-terms of squared Euclidean distance. The structure of deep autoencoder (as displayed in Fig.~\ref{fig_deep_autoencoder}) is such that                
\begin{IEEEeqnarray*}{rCl}
y^{l} & = & \mathcal{G}_{l}(y^{l-1}), \\
& = & \left[\begin{IEEEeqnarraybox*}[][c]{,c/c/c,}  \mathcal{F}_1^l (P^l y^{l-1}) & \cdots &  \mathcal{F}_p^l ( P^l y^{l-1})
 \end{IEEEeqnarraybox*} \right]^T
\end{IEEEeqnarray*}   
where $y^0 = y$, $P^l \in \mathbb{R}^{n_l \times p}$ is a matrix with $n_l \in \{1, \cdots, p\}$ such that $n_1 \geq n_2 \geq \cdots \geq n_L$, and $\mathcal{F}_j^l(\cdot)$ is a Student-t membership-mapping.
\end{definition}
\section{Variational Conditionally Deep Membership-Mapping Autoencoders}\label{sec_3}
This section is dedicated to the variational learning of CDMMAs. Since CDMMA consists of layers of membership-mappings, a solution to the variational learning of membership-mappings is provided in Section~\ref{subsec_variational_learning_membership_mapping}. The solution leads to the design of a learning algorithm for CDMMA in Section~\ref{subsec_738500.4749}. To deal with the big data, a wide CDMMA is presented in Section~\ref{subsec_738500.4831}. The autoencoder based classification is considered in Section~\ref{subsec_738500.4864}.            
\subsection{Variational Learning of Membership-Mappings}\label{subsec_variational_learning_membership_mapping}
\subsubsection{A Modeling Scenario}
Given a dataset $\{(x^i,y^i)\;|\; x^i \in \mathbb{R}^n,\;y^i \in \mathbb{R}^p,\; i \in \{1,\cdots,N \} \}$, it is assumed that there exist zero-mean Student-t membership-mappings $\mathcal{F}_1, \cdots, \mathcal{F}_p \in \mathbb{F}(\mathbb{R}^n)$ such that
\begin{IEEEeqnarray}{rCl}
\label{eq_738118.641846} y^i &\approx & \left[\begin{IEEEeqnarraybox*}[][c]{,c/c/c,} \mathcal{F}_1(x^i)  & \cdots & \mathcal{F}_p(x^i) \end{IEEEeqnarraybox*} \right]^T.
\end{IEEEeqnarray} 
\subsubsection{Disturbances and Auxiliary Inducing Points}
For $ j \in \{1,2,\cdots,p\}$, define
\begin{IEEEeqnarray}{rCl}
\label{eq_y_j_vec2000} \mathrm{y}_j & = & \left[\begin{IEEEeqnarraybox*}[][c]{,c/c/c,}y_j^1 & \cdots &y_j^N\end{IEEEeqnarraybox*} \right]^T \in \mathbb{R}^N\\
\label{eq_f_j_vec} \mathrm{f}_j & = & \left[\begin{IEEEeqnarraybox*}[][c]{,c/c/c,}\mathcal{F}_j(x^1) & \cdots & \mathcal{F}_j(x^N)\end{IEEEeqnarraybox*} \right]^T \in \mathbb{R}^N 
 \end{IEEEeqnarray}  
where $y_j^i$ denotes the $j-$th element of $y^i$. The vectors $ \mathrm{y}_j $ and $ \mathrm{f}_j$ will be subsequently referred to as \emph{data} and \emph{output} of membership-mappings, respectively. The difference between data and membership-mappings' outputs will be referred to as \emph{disturbance} and denoted by $\mathrm{v}_j$, i.e., 
\begin{IEEEeqnarray}{rCl}
\label{eq_v_j_vec} \mathrm{v}_j & = & \mathrm{y}_j  -  \mathrm{f}_j. 
 \end{IEEEeqnarray} 
A set of auxiliary inducing points, $\mathrm{a}  =  \{a^{m} \in \mathbb{R}^{n}\; | \; m \in\{1,\cdots,M\} \}$, is introduced. The membership-mappings' output values at auxiliary inducing input points are collected in a vector defined as
\begin{IEEEeqnarray}{rCl}
\label{738018.270413}\mathrm{u}_j =  \left[\begin{IEEEeqnarraybox*}[][c]{,c/c/c,}\mathcal{F}_j(a^{1}) & \cdots & \mathcal{F}_j(a^{M})\end{IEEEeqnarraybox*} \right]^T \in \mathbb{R}^M.
\end{IEEEeqnarray}
\subsubsection{Membership Functional Representation Approach}
A variable $y \in \mathbf{Y}$ is represented by means of a membership function $\mu_{y}:\mathbf{Y} \rightarrow [0,1]$, where the pair $(\mathbf{Y},\mu_{y})$ constitutes a fuzzy set and $\mu_{y}(\tilde{y})$ is interpreted as the degree to which a point $\tilde{y} \in \mathbf{Y}$ matches an attribute induced by $y \in \mathbf{Y}$. 
\begin{definition}[Disturbance-Model]\label{def_738494.4577}
Disturbance vector $\mathrm{v}_j$ is represented by means of a zero-mean Gaussian membership function as
\begin{IEEEeqnarray}{rCl}
\label{eq_738494.4582}\mu_{\mathrm{v}_j}(\tilde{\mathrm{v}}_j) & = & \exp\left(-0.5\beta \| \tilde{\mathrm{v}}_j \|^2   \right)
\end{IEEEeqnarray}    
where $\beta > 0$ is the precision. 
\end{definition}
\begin{definition}[Representation of Data $\mathrm{y}_j$ for Given Mappings Output $\mathrm{f}_j$]\label{def_738494.462}
Since $\mathrm{y}_j = \mathrm{f}_j + \mathrm{v}_j$, it follows from~(\ref{eq_738494.4582}) that $\mathrm{y}_j$, for given $\mathrm{f}_j$, is represented by means of a membership function, $\mu_{\mathrm{y}_j;\mathrm{f}_j}:\mathbb{R}^N \rightarrow [0,1]$, as
\begin{IEEEeqnarray}{rCl}
\label{eq_membership1002} \mu_{\mathrm{y}_j;\mathrm{f}_j}(\tilde{\mathrm{y}}_j) & = & \exp\left(- 0.5 \beta \|  \tilde{\mathrm{y}}_j -  \mathrm{f}_j \|^2 \right).
\end{IEEEeqnarray}    
\end{definition}
\begin{definition}[Representation of Mappings Output $\mathrm{f}_j$ Based on Interpolation]\label{def_738117.592248}
$\mathrm{f}_j$, based upon an interpolation on the auxiliary-outputs $\mathrm{u}_j $, is represented by means of a membership function, $\mu_{\mathrm{f}_j;\mathrm{u}_j}:\mathbb{R}^N \rightarrow [0,1]$, as
\begin{IEEEeqnarray}{rCl}
\label{eq_pf_1001_student_t}  \mu_{\mathrm{f}_j;\mathrm{u}_j}(\tilde{\mathrm{f}}_j)  & = & \left( 1 + \frac{1}{\nu + M - 2} (\tilde{\mathrm{f}}_j -  \bar{m}_{\mathrm{f}_j})^T  \left( \frac{\nu + (\mathrm{u}_j)^T (K_{\mathrm{a}\mathrm{a}})^{-1} \mathrm{u}_j  - 2}{\nu + M - 2} \bar{K}_{\mathrm{x}\mathrm{x}} \right)^{-1}(\tilde{\mathrm{f}}_j -  \bar{m}_{\mathrm{f}_j}) \right)^{-\frac{\nu+M+N}{2}} \\
\label{eq_pf_1002}  \bar{m}_{\mathrm{f}_j} & = &  K_{\mathrm{x}\mathrm{a}} (K_{\mathrm{a}\mathrm{a}})^{-1}  \mathrm{u}_j   \\
\label{eq_pf_1003} \bar{K}_{\mathrm{x}\mathrm{x}} & = & K_{\mathrm{x}\mathrm{x}} - K_{\mathrm{x}\mathrm{a}} (K_{\mathrm{a}\mathrm{a}})^{-1} K_{\mathrm{x}\mathrm{a}}^T.
\end{IEEEeqnarray}       
\end{definition}
\begin{definition}[Representation of Data $\mathrm{y}_j$ for Fixed Auxiliary-Outputs $\mathrm{u}_j$]
$\mathrm{y}_j$, for given $\mathrm{u}_j$, is represented by means of a membership function, $\mu_{\mathrm{y}_j;\mathrm{u}_j}:\mathbb{R}^N \rightarrow [0,1]$, as
\begin{IEEEeqnarray}{rCl}
\label{eq_738494.4689}   \mu_{\mathrm{y}_j;\mathrm{u}_j}(\tilde{\mathrm{y}}_j)  & \propto  & \exp \left( \left< \log ( \mu_{\mathrm{y}_j;\mathrm{f}_j}(\tilde{\mathrm{y}}_j) )  \right >_{\mu_{\mathrm{f}_j;\mathrm{u}_j}} \right) \IEEEeqnarraynumspace
\end{IEEEeqnarray} 
where $\mu_{\mathrm{y}_j;\mathrm{f}_j}$ is given by (\ref{eq_membership1002}), $\mu_{\mathrm{f}_j;\mathrm{u}_j}$ is defined as in (\ref{eq_pf_1001_student_t}), and $<\cdot>_{\cdot}$ is the averaging operation as defined in (\ref{eq_738118.427179}). Thus, $\mu_{\mathrm{y}_j;\mathrm{u}_j}$ is obtained from $\log( \mu_{\mathrm{y}_j;\mathrm{f}_j})$ after averaging out the variables  $\mathrm{f}_j$ using its membership function. It is shown in \ref{appendix_1} that
\begin{IEEEeqnarray}{rCl}
\label{eq_log_membership2}  \mu_{\mathrm{y}_j;\mathrm{u}_j}(\tilde{\mathrm{y}}_j) & \propto & \exp \left(  - 0.5\beta \| \tilde{\mathrm{y}}_j\|^2  + (\mathrm{u}_j)^T \hat{K}_{\mathrm{u}_j}^{-1} \hat{m}_{\mathrm{u}_j }(\tilde{\mathrm{y}}_j)  - 0.5 (\mathrm{u}_j)^T \hat{K}_{\mathrm{u}_j}^{-1} \mathrm{u}_j + 0.5 (\mathrm{u}_j)^T  (K_{\mathrm{a}\mathrm{a}})^{-1}  \mathrm{u}_j +  \{/ (\tilde{\mathrm{y}}_j,\mathrm{u}_j)\} \right) \IEEEeqnarraynumspace
\end{IEEEeqnarray}  
where $\hat{K}_{\mathrm{u}_j}$, $\hat{m}_{\mathrm{u}_j}(\tilde{\mathrm{y}}_j)$ are given by (\ref{eq_hat_K_u_1000}), (\ref{eq_hat_m_u_1000}) respectively, and $ \{/ (\tilde{\mathrm{y}}_j,\mathrm{u}_j)\}$ represents all those terms which are independent of both $\tilde{\mathrm{y}}_j$ and $\mathrm{u}_j$. The constant of proportionality in (\ref{eq_log_membership2}) is chosen to exclude $ (\tilde{\mathrm{y}}_j,\mathrm{u}_j)-$independent terms in the expression for $\mu_{\mathrm{y}_j;\mathrm{u}_j}$, i.e., 
\begin{IEEEeqnarray}{rCl}
\label{eq_scch_1}  \mu_{\mathrm{y}_j;\mathrm{u}_j}(\tilde{\mathrm{y}}_j)  & = & \exp\left( - 0.5\beta \| \tilde{\mathrm{y}}_j\|^2  + (\mathrm{u}_j)^T \hat{K}_{\mathrm{u}_j}^{-1} \hat{m}_{\mathrm{u}_j }(\tilde{\mathrm{y}}_j)  - 0.5 (\mathrm{u}_j)^T \hat{K}_{\mathrm{u}_j}^{-1} \mathrm{u}_j + 0.5 (\mathrm{u}_j)^T  (K_{\mathrm{a}\mathrm{a}})^{-1}  \mathrm{u}_j \right). \IEEEeqnarraynumspace
\end{IEEEeqnarray}          
\end{definition}
\begin{definition}[Data-Model]\label{def_avg_avg_membership}
$\mathrm{y}_j$ is represented by means of a membership function, $\mu_{\mathrm{y}_j} : \mathbb{R}^N \rightarrow [0,1]$, as 
\begin{IEEEeqnarray}{rCl}
\label{eq_scch_2}  \mu_{\mathrm{y}_j}(\tilde{\mathrm{y}}_j) & \propto & \exp \left( \left<  \log( \mu_{\mathrm{y}_j;\mathrm{u}_j} (\tilde{\mathrm{y}}_j) )\right>_{\mu_{\mathrm{u}_j}} \right)
\end{IEEEeqnarray} 
where $ \mu_{\mathrm{y}_j;\mathrm{u}_j}$ is given by (\ref{eq_scch_1}) and $\mu_{\mathrm{u}_j}:  \mathbb{R}^M \rightarrow [0,1]$ is a membership function representing $\mathrm{u}_j$. Thus, $\mu_{\mathrm{y}_j}$ is obtained from $\log(\mu_{\mathrm{y}_j;\mathrm{u}_j})$ after averaging out the auxiliary-outputs $\mathrm{u}_j$ using membership function $\mu_{\mathrm{u}_j}$. 
\end{definition}   
\subsubsection{Variational Optimization of Data-Model}
The data model (\ref{eq_scch_2}) involves the membership function $\mu_{\mathrm{u}_j}$. To determine $\mu_{\mathrm{u}_j}$ for a given $\mathrm{y}_j$, $\log(\mu_{\mathrm{y}_j}(\mathrm{y}_j))$ is maximized w.r.t. $\mu_{\mathrm{u}_j}$ around an initial guess. The zero-mean Gaussian membership function with covariance as equal to $K_{\mathrm{a}\mathrm{a}}$ is taken as the initial guess. It follows from (\ref{eq_scch_2}) that maximization of $\log(\mu_{\mathrm{y}_j}(\mathrm{y}_j))$ is equivalent to the maximization of $\left<  \log( \mu_{\mathrm{y}_j;\mathrm{u}_j}(\mathrm{y}_j) )\right>_{\mu_{\mathrm{u}_j}}$.  
\begin{result}\label{result_optimal_u}
The solution of following maximization problem:
\begin{IEEEeqnarray}{rCl}
\mu^*_{\mathrm{u}_j}   & = &  \arg \: \max_{\displaystyle \mu_{\mathrm{u}_j}} \: \left[ \left<  \log( \mu_{\mathrm{y}_j;\mathrm{u}_j}(\mathrm{y}_j) )\right>_{\mu_{\mathrm{u}_j}} - \left< \log(\frac{ \mu_{\mathrm{u}_j}(\mathrm{u}_j)}{\exp\left(-0.5 (\mathrm{u}_j)^T (K_{\mathrm{a}\mathrm{a}})^{-1} \mathrm{u}_j  \right)} ) \right>_{ \mu_{\mathrm{u}_j}} \right] \IEEEeqnarraynumspace
\end{IEEEeqnarray} 
under the fixed integral constraint: 
\begin{IEEEeqnarray}{rCl}
\int_{\mathbb{R}^M} \mu_{\mathrm{u}_j}  \, \dd \lambda^M   =  C_{\mathrm{u}_j} > 0
\end{IEEEeqnarray} 
 where the value of $C_{\mathrm{u}_j}$ is so chosen such that the maximum possible values of $\mu^*_{\mathrm{u}_j} $ remain as equal to unity, is given as
\begin{IEEEeqnarray}{rCl}
\label{eq_q_u_vec_optimal} \mu^*_{\mathrm{u}_j}(\mathrm{u}_j)  & = & \exp\left(- 0.5 \left(\mathrm{u}_j - \hat{m}_{\mathrm{u}_j}(\mathrm{y}_j)\right)^T \hat{K}_{\mathrm{u}_j}^{-1}\left(\mathrm{u}_j - \hat{m}_{\mathrm{u}_j}(\mathrm{y}_j)\right) \right)  \IEEEeqnarraynumspace
\end{IEEEeqnarray}
where $\hat{K}_{\mathrm{u}_j}$ and $\hat{m}_{\mathrm{u}_j}$ are given by (\ref{eq_hat_K_u_1000}) and (\ref{eq_hat_m_u_1000}) respectively. The solution of the optimization problem results in  
\begin{IEEEeqnarray}{rCl}
\nonumber \mu_{\mathrm{y}_j}(\tilde{\mathrm{y}}_j)  & \propto & \exp \left( - 0.5\beta\left\{\| \tilde{\mathrm{y}}_j\|^2 - 2 \left(\hat{m}_{\mathrm{u}_j }(\mathrm{y}_j)\right)^T (K_{\mathrm{a}\mathrm{a}})^{-1} (K_{\mathrm{x}\mathrm{a}})^T \tilde{\mathrm{y}}_j +  \left(\hat{m}_{\mathrm{u}_j }(\mathrm{y}_j)\right)^T(K_{\mathrm{a}\mathrm{a}})^{-1}  K_{\mathrm{x}\mathrm{a}}^T K_{\mathrm{x}\mathrm{a}} (K_{\mathrm{a}\mathrm{a}})^{-1}\hat{m}_{\mathrm{u}_j }(\mathrm{y}_j)  \right. \right. \\
 \label{eq_satguru_8}  && \left.  \left.  {+}\: \left(\hat{m}_{\mathrm{u}_j }(\mathrm{y}_j)\right)^T\frac{Tr(K_{\mathrm{x}\mathrm{x}}) - Tr((K_{\mathrm{a}\mathrm{a}})^{-1}  K_{\mathrm{x}\mathrm{a}}^T K_{\mathrm{x}\mathrm{a}} )}{\nu + M - 2} (K_{\mathrm{a}\mathrm{a}})^{-1}\hat{m}_{\mathrm{u}_j }(\mathrm{y}_j)  \right \}  +  \{/ (\mathrm{y}_j,\tilde{\mathrm{y}}_j)\} \right)
\end{IEEEeqnarray} 
where $\{/ (\mathrm{y}_j,\tilde{\mathrm{y}}_j)\}$ represents all $(\mathrm{y}_j,\tilde{\mathrm{y}}_j)-$independent terms.
\end{result}
\begin{IEEEproof}
The proof is provided in \ref{appendix_2}.
\end{IEEEproof}
The constant of proportionality in (\ref{eq_satguru_8}) is chosen to exclude $(\mathrm{y}_j,\tilde{\mathrm{y}}_j)-$independent terms resulting in
\begin{IEEEeqnarray}{rCl}
\nonumber \mu_{\mathrm{y}_j}(\tilde{\mathrm{y}}_j) & = &  \exp \left( - 0.5\beta\left\{\| \tilde{\mathrm{y}}_j\|^2 - 2 \left(\hat{m}_{\mathrm{u}_j }(\mathrm{y}_j)\right)^T (K_{\mathrm{a}\mathrm{a}})^{-1} (K_{\mathrm{x}\mathrm{a}})^T \tilde{\mathrm{y}}_j +  \left(\hat{m}_{\mathrm{u}_j }(\mathrm{y}_j)\right)^T(K_{\mathrm{a}\mathrm{a}})^{-1}  K_{\mathrm{x}\mathrm{a}}^T K_{\mathrm{x}\mathrm{a}} (K_{\mathrm{a}\mathrm{a}})^{-1}\hat{m}_{\mathrm{u}_j }(\mathrm{y}_j)  \right. \right. \\
 \label{eq_738499.7546}  && \left.  \left.  {+}\: \left(\hat{m}_{\mathrm{u}_j }(\mathrm{y}_j)\right)^T\frac{Tr(K_{\mathrm{x}\mathrm{x}}) - Tr((K_{\mathrm{a}\mathrm{a}})^{-1}  K_{\mathrm{x}\mathrm{a}}^T K_{\mathrm{x}\mathrm{a}} )}{\nu + M - 2} (K_{\mathrm{a}\mathrm{a}})^{-1}\hat{m}_{\mathrm{u}_j }(\mathrm{y}_j)  \right \} \right).
\end{IEEEeqnarray}
\subsubsection{Membership-Mapping Output}
 \begin{definition}[Averaged Estimation of Membership-Mapping Output]\label{def_satguru_1}
$\mathcal{F}_j(x^{i})$ (which is the $i-$th element of vector $\mathrm{f}_j$~(\ref{eq_f_j_vec})) can be estimated as
 \begin{IEEEeqnarray}{rCl} 
\widehat{ \mathcal{F}_j(x^{i})} & := &  \left<  \left< (\mathrm{f}_j)_i \right>_{\mu_{\mathrm{f}_j;\mathrm{u}_j } }  \right>_{ \mu^*_{\mathrm{u}_j}}
\end{IEEEeqnarray} 
where $(\mathrm{f}_j)_i$ denotes the $i-$th element of $\mathrm{f}_j$, $\mu_{\mathrm{f}_j;\mathrm{u}_j } $ is defined as in~(\ref{eq_pf_1001_student_t}), and $\mu^*_{\mathrm{u}_j}$ is the optimal membership function (\ref{eq_q_u_vec_optimal}) representing $\mathrm{u}_j$. That is, $\mathcal{F}_j(x^{i})$, being a function of $\mathrm{u}_j$, is averaged over $\mathrm{u}_j$ for an estimation. Let $G(x) \in \mathbb{R}^{1 \times M}$ be a vector-valued function defined as
\begin{IEEEeqnarray}{rCl}
\label{eq_738495.5497} G(x)& := &  \left[\begin{IEEEeqnarraybox*}[][c]{,c/c/c,}kr(x,a^{1}) & \cdots & kr(x,a^{M}) \end{IEEEeqnarraybox*} \right]
 \end{IEEEeqnarray} 
where $kr: \mathbb{R}^n \times \mathbb{R}^n \rightarrow \mathbb{R}$ is defined as in (\ref{eq_membership1003_3}). It is shown in \ref{appendix_738242.626259} that
\begin{IEEEeqnarray}{rCl}
\label{eq_satguru_12}  \widehat{ \mathcal{F}_j(x^{i})}   &=& \left(G(x^i) \right) \left( K_{\mathrm{x}\mathrm{a}}^T K_{\mathrm{x}\mathrm{a}}   + \frac{Tr(K_{\mathrm{x}\mathrm{x}}) - Tr((K_{\mathrm{a}\mathrm{a}})^{-1}  K_{\mathrm{x}\mathrm{a}}^T K_{\mathrm{x}\mathrm{a}} )}{\nu+ M - 2}  K_{\mathrm{a}\mathrm{a}}   +    \frac{K_{\mathrm{a}\mathrm{a}}}{\beta }\right)^{-1}  (K_{\mathrm{x}\mathrm{a}} )^T \mathrm{y}_j . \IEEEeqnarraynumspace
 \end{IEEEeqnarray} 
\end{definition}
Let $\alpha = \left[\begin{IEEEeqnarraybox*}[][c]{,c/c/c,}  \alpha_1 & \cdots & \alpha_p
 \end{IEEEeqnarraybox*} \right] \in \mathbb{R}^{M \times p}$ be a matrix with its $j-$th column defined as 
\begin{IEEEeqnarray}{rCl}
\label{eq_vector_alpha}  \alpha_j & := & \left( K_{\mathrm{x}\mathrm{a}}^T K_{\mathrm{x}\mathrm{a}}   + \frac{Tr(K_{\mathrm{x}\mathrm{x}}) - Tr((K_{\mathrm{a}\mathrm{a}})^{-1}  K_{\mathrm{x}\mathrm{a}}^T K_{\mathrm{x}\mathrm{a}} )}{\nu+ M - 2}  K_{\mathrm{a}\mathrm{a}}   +    \frac{K_{\mathrm{a}\mathrm{a}}}{\beta }\right)^{-1}  (K_{\mathrm{x}\mathrm{a}} )^T \mathrm{y}_j  
  \end{IEEEeqnarray}
  so that $\widehat{ \mathcal{F}_j(x^{i})}$ could be expressed as    
\begin{IEEEeqnarray}{rCl}
\label{eq_final_layer_output} \widehat{ \mathcal{F}_j(x^{i})} & = & \left(G(x^i) \right) \alpha_j.
  \end{IEEEeqnarray}   
\subsubsection{Choice of Parameters}
The analytically derived data model (\ref{eq_738499.7546}) involves several parameters which are suggested to be chosen as follows:  
\paragraph{Auxiliary inducing points:}
The auxiliary inducing points are suggested to be chosen as the cluster centroids: 
\begin{IEEEeqnarray}{rCl}
\label{eq_738496.4692}\mathrm{a} = \{ a^{m}\}_{m=1}^M  =  cluster\_centroid(  \{x^i\}_{i=1}^N, M) 
 \end{IEEEeqnarray} 
where $cluster\_centroid(  \{ x^i \}_{i=1}^N,M)$ represents the k-means clustering on $ \{ x^i \}_{i=1}^N$. 
\paragraph{Degrees of freedom:}
The degrees of freedom associated to the Student-t membership-mapping $\nu \in \mathbb{R}_{+} \setminus [0,2]$ is chosen as 
\begin{IEEEeqnarray}{rCl}
\label{eq_738496.4701}\nu & = & 2.1
 \end{IEEEeqnarray} 
\paragraph{Parameters $(w_1,\cdots,w_n)$:}
The parameters $(w_1,\cdots,w_n)$ for kernel function~(\ref{eq_membership1003_3}) are chosen such that $w_{k}$ (for $k\in \{1,2,\cdots,n\}$) is given as
\begin{IEEEeqnarray}{rCl}
\label{eq_738496.4698}w_k & = & \left(\max_{1 \leq i \leq N}\left(x^i_k\right) - \min_{1 \leq i \leq N}\left(x^i_k\right)\right)^{-2}
 \end{IEEEeqnarray} 
where $x^i_k$ is the $k-$th element of vector $x^i \in \mathbb{R}^n$.
\paragraph{Parameters $M$ and $\sigma^2$:}
Define a scalar-valued function:
\begin{IEEEeqnarray}{rCl}
\tau(M,\sigma^2) & := & \frac{Tr(K_{\mathrm{x}\mathrm{x}}) - Tr((K_{\mathrm{a}\mathrm{a}})^{-1}  K_{\mathrm{x}\mathrm{a}}^T K_{\mathrm{x}\mathrm{a}} )}{\nu+ M - 2}
 \end{IEEEeqnarray} 
where $\mathrm{a}$ is given by (\ref{eq_738496.4692}), $\nu$ is given by (\ref{eq_738496.4701}), and parameters $(w_1,\cdots,w_n)$ (which are required to evaluate the kernel function for computing matrices $K_{\mathrm{x}\mathrm{x}}$, $K_{\mathrm{a}\mathrm{a}}$, and $K_{\mathrm{x}\mathrm{a}}$) are given by (\ref{eq_738496.4698}). It follows from the kernel function definition~(\ref{eq_membership1003_3}) that
\begin{IEEEeqnarray}{rCl}
\label{eq_738497.3805} \tau(M,\sigma^2) & = & \sigma^2\tau(M,1).
 \end{IEEEeqnarray}  
It is further observed from (\ref{eq_satguru_12}) that a higher value of $\tau$ corresponds to a larger level of data smoothing. Therefore, we consider a criterion for choosing $M$ and $\sigma^2$ such that the data smoothing level (i.e. the value of $\tau$) should match the variance in the data. In particularly, we pose the requirement that $\tau$ should be at least as large as the data variance (averaged over dimensions), i.e.,
\begin{IEEEeqnarray}{rCl}
\label{eq_738496.6594} \tau(M,\sigma^2) & \geq & \frac{1}{p} \sum_{j=1}^p \text{var}\left(y_j^1,\cdots,y_j^N\right).
 \end{IEEEeqnarray}     
Using (\ref{eq_738497.3805}), the inequality (\ref{eq_738496.6594}) can be rewritten as
\begin{IEEEeqnarray}{rCl}
 \label{eq_738497.3839} \sigma^2 & \geq & \frac{1}{\tau(M,1)} \frac{1}{p} \sum_{j=1}^p \text{var}\left(y_j^1,\cdots,y_j^N\right).
 \end{IEEEeqnarray}  
The number of auxiliary inducing points $M$ and the parameter $\sigma^2$ for kernel function~(\ref{eq_membership1003_3}) are so determined that the inequality (\ref{eq_738497.3839}) holds. This can be done via
\begin{enumerate}
\item choosing sufficiently low value of $M$ ensuring that $\tau(M,1)$ remains larger than a small positive value,   
\item choosing $\sigma^2$ to satisfy the inequality (\ref{eq_738497.3839}).
\end{enumerate}
\paragraph{Precision of the disturbance model:}
The disturbance precision value $\beta$ is iteratively estimated as the inverse of the mean squared error between data and membership-mappings outputs. That is, 
\begin{IEEEeqnarray}{rCl}
\label{eq_738497.4473} \frac{1}{\beta} & = & \frac{1}{pN}\sum_{j=1}^p \sum_{i=1}^N \left |y_j^i -  \widehat{ \mathcal{F}_j(x^{i})} \right |^2
 \end{IEEEeqnarray} 
where $\widehat{ \mathcal{F}_j(x^{i})}$ is the estimated membership-mapping output given as in~(\ref{eq_final_layer_output}). 
\subsubsection{Learning Algorithm and Prediction}
Algorithm~\ref{algorithm_basic_learning} is suggested for the variational learning of membership-mappings. The functionality of Algorithm~\ref{algorithm_basic_learning} is as follows.
\begin{enumerate}
\item The loop between step~4 and step~7 ensures, via gradually decreasing the number of auxiliary points by a factor of 0.9, that $\tau(M,1)$ is positive with value larger than $\kappa$. 
\item The positive value of $\tau(M,1)$ allows steps~9~to~13 to ensure that the inequality (\ref{eq_738497.3839}) remains satisfied and thus the level of data smoothing by membership-mappings remains related to the data variance. 
\item The loop between step~16 and step~19 iteratively estimates the parameters $\alpha$ and $\beta$.   
\end{enumerate} 
\begin{algorithm}
\caption{Variational learning of the membership-mappings}
\label{algorithm_basic_learning}
\begin{algorithmic}[1]
\REQUIRE  Dataset $\left\{ (x^i,y^i) \; | \; x^i \in \mathbb{R}^n,\; y^i \in \mathbb{R}^p,\; i \in \{1,\cdots,N \} \right \}$ and maximum possible number of auxiliary points $M_{max} \in \mathbb{Z}_+$ with $M_{max} \leq N$.  
\STATE Choose $\nu$ and $w = (w_1,\cdots,w_n)$ as in (\ref{eq_738496.4701}) and (\ref{eq_738496.4698}) respectively.  
\STATE Choose a small positive value $\kappa = 10^{-1}$. 
\STATE Set iteration count $it = 0$ and $M|_0 = M_{max}$.
\WHILE{$\tau(M|_{it},1) < \kappa$}
\STATE $M|_{it+1} = \lceil 0.9M|_{it} \rceil$
\STATE $it \leftarrow it + 1$
\ENDWHILE
\STATE Set $M = M|_{it}$.
\IF{$\tau(M,1) \geq  \frac{1}{p} \sum_{j=1}^p \text{var}\left(y_j^1,\cdots,y_j^N\right)$}
\STATE $\sigma^2 = 1$
\ELSE 
\STATE $\sigma^2 = \frac{1}{\tau(M,1)} \frac{1}{p} \sum_{j=1}^p \text{var}\left(y_j^1,\cdots,y_j^N\right)$
\ENDIF
\STATE Compute $\mathrm{a} = \{ a^{m}\}_{m=1}^M$ using (\ref{eq_738496.4692}), $K_{\mathrm{x}\mathrm{x}}$ using (\ref{738026.844153}), $K_{\mathrm{a}\mathrm{a}}$ using (\ref{eq_membership1004_2}), and $K_{\mathrm{x}\mathrm{a}}$ using (\ref{eq_738497.4922}).
\STATE Set $\beta = 1$.
 \REPEAT
\STATE Compute $\alpha$ using (\ref{eq_vector_alpha}).
\STATE Update the value of $\beta$ using (\ref{eq_738497.4473}).
 \UNTIL{($\beta$ nearly converges)}
\STATE Compute $\alpha$ using (\ref{eq_vector_alpha}).
 \RETURN the parameters set $\mathbb{M} = \{\alpha, \mathrm{a}, M,\sigma,w\}$.
\end{algorithmic} 
\end{algorithm} 
\begin{definition}[Membership-Mappings Prediction]
Given the parameters set $\mathbb{M} = \{\alpha, \mathrm{a}, M,\sigma,w\}$ returned by Algorithm~\ref{algorithm_basic_learning}, the learned membership-mappings could be used to predict output corresponding to any arbitrary input data point $x \in \mathbb{R}^n$ as
\begin{IEEEeqnarray}{rCl}
\hat{y}(x;\mathbb{M}) & = & \left[\begin{IEEEeqnarraybox*}[][c]{,c/c/c,} \widehat{ \mathcal{F}_1(x)} & \cdots & \widehat{ \mathcal{F}_p(x)}\end{IEEEeqnarraybox*} \right]^T
 \end{IEEEeqnarray}  
where $\widehat{ \mathcal{F}_j(x)}$, defined as in (\ref{eq_final_layer_output}), is the estimated output of $j-$th membership-mapping. It follows from (\ref{eq_final_layer_output}) that  
\begin{IEEEeqnarray}{rCl}
\label{eq_738124.770095}\hat{y}(x;\mathbb{M}) & = & \alpha^T(G(x))^T
\end{IEEEeqnarray}
where $G(\cdot) \in \mathbb{R}^{1 \times M}$ is a vector-valued function~(\ref{eq_738495.5497}). 
\end{definition}
\subsection{Algorithm for Variational Learning of Conditionally Deep Membership-Mapping Autoencoders}\label{subsec_738500.4749} 
Since CDMMA consists of layers of membership-mappings, Algorithm~\ref{algorithm_basic_learning} could be directly applied for the variational learning of individual layers. Given a set of $N$ samples $\{y^1,\cdots,y^N \}$, following~\cite{10.1007/978-3-030-87101-7_14}, Algorithm~\ref{algorithm_DSFMA} is stated for the variational learning of CDMMA.       
\begin{algorithm}
\caption{Variational learning of CDMMA}
\label{algorithm_DSFMA}
\begin{algorithmic}[1]
\REQUIRE Data set $\mathbf{Y} = \left\{ y^i \in \mathbb{R}^p \; | \; i \in \{1,\cdots,N \} \right \}$; the subspace dimension $n \in \{1,2,\cdots,p \}$; maximum number of auxiliary points $M_{max} \in \mathbb{Z}_+$ with $M_{max} \leq N$; the number of layers $L \in \mathbb{Z}_{+}$.
\FOR{$l=1$ to $L$}
\STATE Set subspace dimension associated to $l-$th layer as $n_l = \max(n - l + 1,1)$.
\STATE Define $P^l \in \mathbb{R}^{n_l \times p}$ such that $i-$th row of $P^l$ is equal to transpose of eigenvector corresponding to $i-$th largest eigenvalue of sample covariance matrix of data set $\mathbf{Y} $. 
\STATE Define a latent variable $x^{l,i} \in \mathbb{R}^{n_l}$, for $i \in \{1,\cdots,N \}$, as
 \begin{IEEEeqnarray}{rCl}
\label{eq_x_l_i}x^{l,i} &:=& \left\{ \,
    \begin{IEEEeqnarraybox}[][c]{l?s}
      \IEEEstrut
      P^ly^i & if $l=1$, \\
     P^l \hat{y}^{l-1}(x^{l-1,i};\mathbb{M}^{l-1}) & if $l > 1$
      \IEEEstrut
    \end{IEEEeqnarraybox}
\right.  \IEEEeqnarraynumspace
\end{IEEEeqnarray}   
where $\hat{y}^{l-1}$ is the estimated output of the $(l-1)-$th layer computed using (\ref{eq_738124.770095}) for the parameters set $\mathbb{M}^{l-1} = \{\alpha^{l-1}, \mathrm{a}^{l-1}, M^{l-1}, \sigma^{l-1},w^{l-1}\}$.  
\STATE Define $M_{max}^l$ as
 \begin{IEEEeqnarray}{rCl}
\label{eq_738499.4927}M_{max}^l &:=& \left\{ \,
    \begin{IEEEeqnarraybox}[][c]{l?s}
      \IEEEstrut
      M_{max} & if $l=1$, \\
     M^{l-1} & if $l > 1$
      \IEEEstrut
    \end{IEEEeqnarraybox}
\right.  \IEEEeqnarraynumspace
\end{IEEEeqnarray} 
\STATE Compute parameters set $\mathbb{M}^l = \{\alpha^{l}, \mathrm{a}^{l}, M^{l}, \sigma^{l},w^{l}\}$, characterizing the membership-mappings associated to $l-$th layer, using Algorithm~\ref{algorithm_basic_learning} on data set $\left\{ (x^{l,i},y^i) \; | \;  i \in \{1,\cdots,N \} \right \}$ with maximum possible number of auxiliary points $M_{max}^l$. 
\ENDFOR
 \RETURN the parameters set $\mathcal{M} = \{\{\mathbb{M}^1,\cdots,\mathbb{M}^L\}, \{P^1,\cdots,P^L \} \}$.
\end{algorithmic}
\end{algorithm} 

Algorithm~\ref{algorithm_DSFMA} defines $\{n_1,\cdots,n_L\}$ to be a monotonically decreasing sequence at step~2. This results in CDMMA to discover layers of increasingly abstract data representation with lowest-level data features being modeled by first layer and the highest-level by end layer.    
\begin{figure}[bt]
\centering
\includegraphics[width = \textwidth]{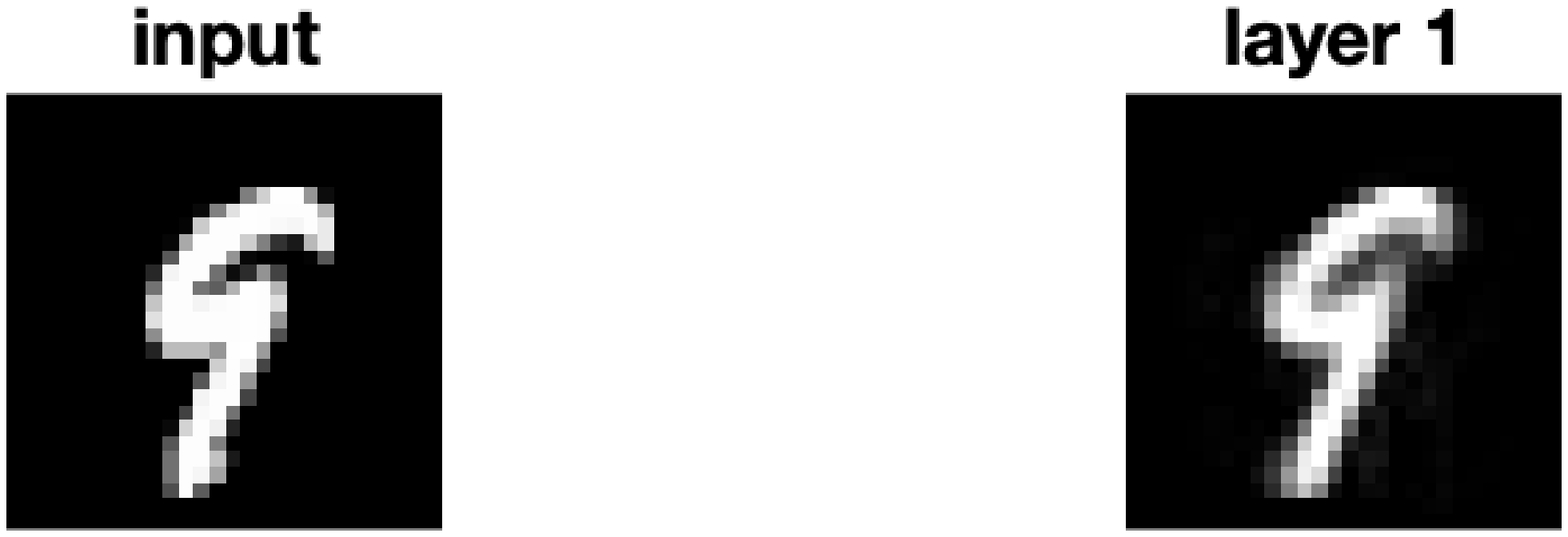}
\caption{A CDMMA was built using Algorithm~\ref{algorithm_DSFMA} (taking $n = 20$; $M_{max} = 500$; $L = 20$) on a dataset consisting of 1000 randomly chosen samples of digit 9 from MNIST digits dataset. Corresponding to the input sample (shown at the extreme left of the figure), the estimated outputs of different layers of CDMMA are displayed.}
\label{fig_deep_learning}
\end{figure}   
Fig.~\ref{fig_deep_learning} illustrates through an example that Algorithm~\ref{algorithm_DSFMA} allows high-dimensional data representation learning at varying abstraction levels across CDMMA's different layers.     
\begin{definition}[CDMMA Filtering]\label{def_DSFMA_filtering}
Given a CDMMA with its parameters being represented by a set $\mathcal{M} = \{\{\mathbb{M}^1,\cdots,\mathbb{M}^L\}, \{P^1,\cdots,P^L \} \}$, the autoencoder can be applied for filtering a given input vector $y \in \mathbb{R}^p$ as follows:   
 \begin{IEEEeqnarray}{rCl}
x^l(y;\mathcal{M}) &=& \left\{ \,
    \begin{IEEEeqnarraybox}[][c]{l?s}
      \IEEEstrut
      P^ly, & $l=1$ \\
      P^l  \hat{y}^{l-1}(x^{l-1};\mathbb{M}^{l-1})&  $l \geq 2$
      \IEEEstrut
    \end{IEEEeqnarraybox}
\right. 
\end{IEEEeqnarray} 
Here, $\hat{y}^{l-1}$ is the output of the $(l-1)-$th layer estimated using (\ref{eq_738124.770095}). Finally, CDMMA's output, $\mathcal{D}(y;\mathcal{M})$, is given as
\begin{IEEEeqnarray}{rCl}
\label{eq_satguru_18}  \widehat{\mathcal{D}}(y;\mathcal{M}) & = &  \hat{y}^{l^*}(x^{l^*};\mathbb{M}^{l^*}) \\
\label{eq_satguru_19} l^*  & = & \arg\;\min_{l \: {\in} \: \{1,\cdots,L \}}\; \|y - \hat{y}^{l}(x^{l};\mathbb{M}^{l}) \|^2.
 \end{IEEEeqnarray} 
\end{definition}
\subsection{Wide Conditionally Deep Membership-Mapping Autoencoder}\label{subsec_738500.4831}
For big datasets, a wide form of conditionally deep autoencoder has been suggested~\cite{10.1007/978-3-030-87101-7_14} where the total data is partitioned into subsets and corresponding to each data-subset a separate CDMMA is learned. The final output is equal to the output of the CDMMA re-constructing the given input vector as good as possible where re-construction error is measured in-terms of squared Euclidean distance.
\begin{definition}[A Wide CDMMA]\label{def_wide_deep_autoencoder}
A wide CDMMA, $\mathcal{WD}:\mathbb{R}^p \rightarrow \mathbb{R}^p$, maps a vector $y \in \mathbb{R}^p$ to $\mathcal{WD}(y) \in \mathbb{R}^p$ through a parallel composition of $S$  ($S \in \mathcal{Z}_+$) number of CDMMAs such that
 \begin{IEEEeqnarray}{rCl}
\label{eq_738125.489500} \mathcal{WD}(y) & = & \mathcal{D}_{s^*}(y)\\
 s^* & = & \arg\;\min_{s \in \{1,2,\cdots,S \}}\; \| y -  \mathcal{D}_s(y)  \|^2,
 \end{IEEEeqnarray}  
where $\mathcal{D}_s(y)$ is the output of $s-$th CDMMA. 
\end{definition}

\begin{figure}[bt]
\centering
\includegraphics[width = \textwidth]{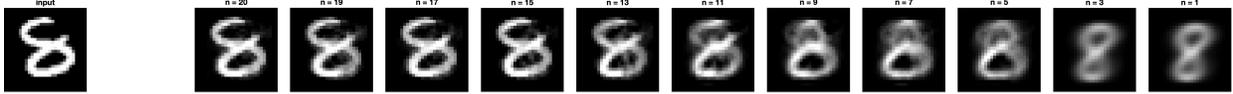}
\caption{On a dataset consisting of 1000 randomly chosen samples of digit 8 from MNIST digits dataset, different wide CDMMAs were built using Algorithm~\ref{algorithm_WDSFMA} choosing $r_{max} = 0.5$, $L = 5$, and $n$ from $\{20,19,17,15, 13,11,9,7,5,3,1\}$. Corresponding to the input sample (shown at the extreme left of the figure), the estimated outputs of different wide CDMMAs (built using different values of $n$) are displayed. It is observed that as $n$ keeps on decreasing, the autoencoder learns increasingly abstract data representation.}
\label{fig_deep_learning_3}
\end{figure}

\begin{figure}[bt]
\centering
\includegraphics[width = \textwidth]{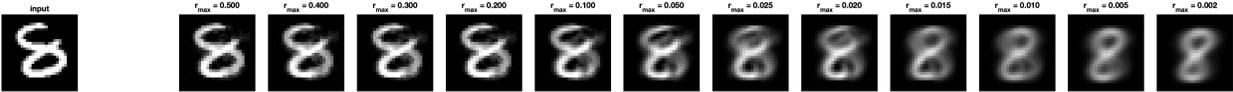}
\caption{On a dataset consisting of 1000 randomly chosen samples of digit 8 from MNIST digits dataset, different wide CDMMAs were built using Algorithm~\ref{algorithm_WDSFMA} choosing $r_{max}$ from $\{0.5,0.4,0.3,0.2,0.1,0.05,0.025,0.02,0.015,0.01,0.005,0.002\}$, $n = 20$, and $L = 5$. Corresponding to the input sample (shown at the extreme left of the figure), the estimated outputs of different wide CDMMAs (built using different $r_{max}$ values) are displayed. It is observed that as the $r_{max}$ value keeps on decreasing, the autoencoder learns increasingly abstract data representation.}
\label{fig_deep_learning_2}
\end{figure}

To formally define an algorithm for the variational learning of wide CDMMA, the ratio of maximum number of auxiliary points to the number of data points is defined:
 \begin{IEEEeqnarray}{rCl}
 r_{max} & = & \frac{M_{max}}{N}.
 \end{IEEEeqnarray} 
Following~\cite{10.1007/978-3-030-87101-7_14}, Algorithm~\ref{algorithm_WDSFMA} is suggested for the variational learning of wide CDMMA.      
\begin{algorithm}
\caption{Variational learning of wide CDMMA}
\label{algorithm_WDSFMA}
\begin{algorithmic}[1]
\REQUIRE  Data set $\mathbf{Y} = \left\{ y^i \in \mathbb{R}^p \; | \; i \in \{1,\cdots,N \} \right \}$; the subspace dimension $n \in \{1,2,\cdots,p\}$; ratio $r_{max} \in (0,1]$; the number of layers $L \in \mathbb{Z}_{+}$.
\STATE Apply k-means clustering to partition $\mathbf{Y} $ into $S$ subsets, $\{\mathbf{Y}^1, \cdots, \mathbf{Y}^S  \}$, where $S = \lceil N/1000 \rceil$. 
\FOR{$s = 1$ to $S$}
\STATE Build a CDMMA, $\mathcal{M}^s$, by applying Algorithm~\ref{algorithm_DSFMA} on $\mathbf{Y}^s$ taking $n$ as the subspace dimension; maximum number of auxiliary points as equal to $r_{max} \times \#\mathbf{Y}^s$ (where $\#\mathbf{Y}^s$ is the number of data points in $\mathbf{Y}^s$); and $L$ as the number of layers.   
\ENDFOR
 \RETURN the parameters set $\mathcal{P} = \{\mathcal{M}^s\}_{s=1}^S$.
\end{algorithmic}
\end{algorithm} 
\begin{definition}[Wide CDMMA Filtering]\label{def_DSFMA_filtering}
Given a wide CDMMA with its parameters being represented by a set $\mathcal{P} = \{\mathcal{M}^s\}_{s=1}^S$, the autoencoder can be applied for filtering a given input vector $y \in \mathbb{R}^p$ as follows:   
  \begin{IEEEeqnarray}{rCl}
 \label{eq_738500.4495}\widehat{\mathcal{WD}}(y;\mathcal{P}) & = &   \widehat{\mathcal{D}}(y;\mathcal{M}^{s^*})\\
 s^* & = & \arg\;\min_{s \in \{1,2,\cdots,S \}}\; \| y -  \widehat{\mathcal{D}}(y;\mathcal{M}^{s})  \|^2,
 \end{IEEEeqnarray}   
where $ \widehat{\mathcal{D}}(y;\mathcal{M}^{s})$ is the output of $s-$th CDMMA estimated using (\ref{eq_satguru_18}).
\end{definition}
Algorithm~\ref{algorithm_WDSFMA} requires choosing the values for subspace dimension $n$ and ratio $r_{max}$. Thus, we demonstrate the effect of $n$ and $r_{max}$ on the abstraction level of data representation through examples in Fig.~\ref{fig_deep_learning_3} and Fig.~\ref{fig_deep_learning_2}.   
\subsection{Classification Applications}\label{subsec_738500.4864}
The autoencoders could be applied for classification via learning data representation for each class through a separate autoencoder. Formally, the classifier is defined as in Definition~\ref{def_classifier} and Algorithm~\ref{algorithm_classification} is stated for the variational learning of classifier.   
\begin{definition}[A Classifier]\label{def_classifier}
A classifier, $\mathcal{C}: \mathbb{R}^p \rightarrow \{1,2,\cdots, C\}$, maps a vector $y \in \mathbb{R}^p$ to $\mathcal{C}(y) \in  \{1,2,\cdots, C\}$ such that
\begin{IEEEeqnarray}{rCl}
\label{eq_satguru_20} \mathcal{C}(y; \{\mathcal{P}_c\}_{c=1}^C)
  & = &  \arg\;\min_{c \:{\in}\: \{1,2,\cdots,C \}}\; \| y -   \widehat{\mathcal{WD}}(y;\mathcal{P}_c) \|^2  \end{IEEEeqnarray} 
where $ \widehat{\mathcal{WD}}(y;\mathcal{P}_c)$, computed using (\ref{eq_738500.4495}), is the output of $c-$th wide CDMMA. The classifier assigns to an input vector the label of that class whose associated autoencoder best reconstructs the input vector.      
 \end{definition}
\begin{algorithm}
\caption{Variational learning of the classifier}
\label{algorithm_classification}
\begin{algorithmic}[1]
\REQUIRE Labeled data set $\mathbf{Y} = \left \{ \mathbf{Y}_c\; | \; \mathbf{Y}_c =  \left \{ y^{i,c} \in \mathbb{R}^p \; | \; i \in \{1,\cdots,N_c \} \right \},\: c \in \{1,\cdots,C \} \right \}$; the subspace dimension $n \in \{1,\cdots,p\}$; ratio $r_{max} \in (0,1]$; the number of layers $L \in \mathbb{Z}_{+}$.
\FOR{$c = 1$ to $C$} 
\STATE Build a wide CDMMA, $\mathcal{P}_c = \{\mathcal{M}^s_c\}_{s=1}^{S_c}$, by applying Algorithm~\ref{algorithm_WDSFMA} on $\mathbf{Y}_c$ for the given $n$, $r_{max}$, and $L$.  
\ENDFOR
 \RETURN the parameters set $ \{ \mathcal{P}_c  \}_{c=1}^C$. 
\end{algorithmic}
\end{algorithm} 
\section{Privacy-Preserving Transferrable Deep Learning}\label{sec_4}
\subsection{An Optimal $(\epsilon,\delta)-$Differentially Private Noise Adding Mechanism}\label{sec_4_1}
This subsection reviews an optimal noise adding mechanism that was derived using an information theoretic approach in~\cite{Kumar/IWCFS2019}. We consider a training dataset consisting of $N$ number of samples with each sample having $p$ number of attributes. Assuming the data as numeric, the dataset can be represented by a matrix, say $\mathrm{Y} \in \mathbb{R}^{p \times N}$. The machine learning algorithms typically train a model using available dataset. A given machine learning algorithm, training a model using data matrix $\mathrm{Y}$, can be represented by a mapping, $\mathcal{A}: \mathbb{R}^{p \times N} \rightarrow \mathbf{M}$, where $\mathbf{M}$ is the model space. That is, for a given dataset $\mathrm{Y}$, the algorithm builds a model $\mathcal{M} \in \mathbf{M}$ such that $\mathcal{M}  =  \mathcal{A}(\mathrm{Y})$. The privacy of data can be preserved via adding a suitable random noise to data matrix before the application of algorithm $\mathcal{A}$ on the dataset. This will result in a private version of algorithm $\mathcal{A}$ which is formally defined by Definition~\ref{definition_private_algorithm}.
\begin{definition}[A Private Algorithm on Data Matrix]\label{definition_private_algorithm}
Let $\mathcal{A}^+ : \mathbb{R}^{p \times N} \rightarrow Range(\mathcal{A}^+)$ be a mapping defined as
\begin{IEEEeqnarray}{rCl}
\mathcal{A}^+\left(\mathrm{Y}\right) & = & \mathcal{A}\left(\mathrm{Y} + \mathrm{V}\right),\;  \mathrm{V} \in \mathbb{R}^{p \times N}
\end{IEEEeqnarray}  
where $\mathrm{V}$ is a random noise matrix with $f_{\mathrm{v}_j^i}(v)$ being the probability density function of its $(j,i)-$th element $\mathrm{v}_j^i$; $\mathrm{v}_j^i$ and $\mathrm{v}_j^{i^{\prime}}$ are independent from each other for $i \neq i^{\prime}$; and $\mathcal{A}: \mathbb{R}^{p \times N} \rightarrow \mathbf{M}$ (where $\mathbf{M}$ is the model space) is a given mapping representing a machine learning algorithm. The range of $\mathcal{A}^+$ is as
 \begin{IEEEeqnarray}{rCl}
 Range(\mathcal{A}^+) & = & \left \{ \mathcal{A}\left(\mathrm{Y} + \mathrm{V}\right)\; | \;  \mathrm{Y} \in \mathbb{R}^{p \times N}, \mathrm{V} \in \mathbb{R}^{p \times N} \right \}. 
\end{IEEEeqnarray}  
\end{definition}  
We intend to protect the algorithm $\mathcal{A}^+$ from an adversary who seeks to gain an information about the data from algorithm's output by perturbing the values in a sample of the dataset. We seek to attain differential privacy for algorithm $\mathcal{A}^+$ against the perturbation in an element of $\mathrm{Y}$, say $(j_0,i_0)-$th element, such that magnitude of the perturbation is upper bounded by a scalar $d$. Following~\cite{9133180}, the $d-$adjacency and $(\epsilon,\delta)-$differential privacy definitions are provided in Definition~\ref{def_adjacency_matrices} and Definition~\ref{def_differential_privacy} respectively.        
\begin{definition}[$d-$Adjacency for Data Matrices]\label{def_adjacency_matrices}
Two matrices $\mathrm{Y},\mathrm{Y}^{\prime} \in \mathbb{R}^{p \times N}$ are $d-$adjacent if for a given $d  \in \mathbb{R}_{+}$, there exist $i_0 \in \{1,2,\cdots,N\}$ and $j_0 \in \{1,2,\cdots,p\}$ such that $\forall i \in \{1,2,\cdots,N\}, j \in \{1,2,\cdots,p\}$,
 \begin{IEEEeqnarray*}{rCl}
\left | \mathrm{y}^i_j - \mathrm{y}^{\prime i}_j \right | & \leq & \left\{ \begin{array}{cc}
d, & \mbox{if $i = i_0,j = j_0$} \\
0, & \mbox{otherwise}
  \end{array} \right.
\end{IEEEeqnarray*}    
where $\mathrm{y}^i_j$ and $\mathrm{y}^{\prime i}_j$ denote the $(j,i)-$th element of $\mathrm{Y}$ and $\mathrm{Y}^{\prime}$ respectively. Thus, $\mathrm{Y}$ and $\mathrm{Y}^{\prime}$ differ by only one element and the magnitude of the difference is upper bounded by $d$. 
\end{definition} 
\begin{definition}[$(\epsilon,\delta)-$Differential Privacy for $\mathcal{A}^+$~\cite{Kumar/IWCFS2019}]\label{def_differential_privacy}
The algorithm $\mathcal{A}^+\left(\mathrm{Y}\right)$ is $(\epsilon,\delta)-$differentially private if
 \begin{IEEEeqnarray}{rCl}
\label{eq_differential_privacy}  Pr\{ \mathcal{A}^+\left(\mathrm{Y}\right) \in \mathcal{O} \} & \leq & \exp(\epsilon) Pr\{ \mathcal{A}^+\left(\mathrm{Y}^{\prime}\right)) \in \mathcal{O} \} + \delta
\end{IEEEeqnarray}     
for any measurable set $\mathcal{O} \subseteq  Range(\mathcal{A}^+) $ and for $d-$adjacent matrices pair $(\mathrm{Y},\mathrm{Y}^{\prime})$.     
\end{definition} 
Intuitively, Definition~\ref{def_differential_privacy} means that changing the value of an element in the training data matrix by an amount upper bounded by $d$ can change the distribution of output of the algorithm $\mathcal{A}^+$ only by a factor of $\exp(\epsilon)$ with probability at least $1-\delta$. Thus, the lower value of $\epsilon$ and $\delta$ lead to a higher amount of privacy. 
\begin{result}[An Optimal $(\epsilon,\delta)-$Differentially Private Noise~\cite{Kumar/IWCFS2019}]\label{result_optimal_noise_epsilon_delta_privacy}
The probability density function of noise, that minimizes the expected noise magnitude together with satisfying the sufficient conditions for $(\epsilon,\delta)-$differential privacy for $\mathcal{A}^+$, is given as
\begin{IEEEeqnarray}{rCl}
\label{eq_optimal_density_epsilon_delta_privacy} f_{\mathrm{v}_j^i}^*(v;\epsilon,\delta,d) &  = & \left \{\begin{array}{cl}  {\delta}\: Dirac\delta(v), & v = 0 \\
 (1- \delta)\frac{\displaystyle \epsilon}{\displaystyle  2 d} \exp(-\frac{\displaystyle  \epsilon}{\displaystyle   d} |v|), & v \in   \mathbb{R} \setminus \{0\}
\end{array} \right.
\end{IEEEeqnarray}  
where $Dirac\delta(v)$ is Dirac delta function satisfying $\int_{-\infty}^{\infty}Dirac\delta(v)\: \dd v = 1$. 
\end{result}
 \begin{IEEEproof}
The proof follows from~\cite{Kumar/IWCFS2019}.
 \end{IEEEproof}
\begin{remark}[Generating Random Samples from $f_{\mathrm{v}_j^i}^*$] \label{remark_sampling}
The method of {\it inverse transform sampling} can be used to generate random samples from cumulative distribution function. The cumulative distribution function of $  f_{\mathrm{v}_j^i}^*$ is given as
 \begin{IEEEeqnarray}{rCl}
F_{\mathrm{v}_j^i}(v;\epsilon,\delta,d) & = & \left \{\begin{array}{ll} \frac{ 1-\delta}{  2} \exp(\frac{\epsilon }{d} v), & v < 0 \\
\frac{ 1 + \delta}{  2}, &  v = 0  \\
1 - \frac{ 1 - \delta}{  2}\exp(-\frac{\epsilon }{d} v), & v > 0
\end{array} \right. 
\end{IEEEeqnarray}      
The inverse cumulative distribution function is given as
 \begin{IEEEeqnarray}{rCl}
\label{eq_inverse_cdf}F_{\mathrm{v}_j^i}^{-1}(t^i_j;\epsilon,\delta,d) & = & \left \{\begin{array}{ll} \frac{d}{\epsilon} \log(\frac{2t^i_j}{1 - \delta}), & t^i_j <  \frac{1- \delta}{2} \\
0, &  t^i_j \in [ \frac{1- \delta}{2}, \frac{1+ \delta}{2}]  \\
 -\frac{d}{\epsilon} \log(\frac{2(1-t^i_j)}{1-\delta}), & t^i_j > \frac{1+\delta}{2}
\end{array} \right.,\; t^i_j \in (0,1).
\end{IEEEeqnarray}   
Thus, via generating random samples from the uniform distribution on $(0,1)$ and using~(\ref{eq_inverse_cdf}), the noise additive mechanism can be implemented.   
\end{remark} 
\begin{algorithm}
\caption{Differentially private approximation of data samples}
\label{algorithm_differential_private_approximation}
\begin{algorithmic}[1]
\REQUIRE  Data set $\mathbf{Y} =  \left \{ y^{i} \in \mathbb{R}^p \; | \; i \in \{1,\cdots,N \} \right \}$; differential privacy parameters: $d  \in \mathbb{R}_{+}$,  $\epsilon  \in \mathbb{R}_{+}$, $\delta \in (0,1)$.
\STATE A differentially private approximation of data samples is provided as
\begin{IEEEeqnarray}{rCl}
 y^{+i}_j & = & y^{i}_j + F_{\mathrm{v}_j^{i}}^{-1}(t^{i}_j;\epsilon,\delta,d),\; t^{i}_j \in (0,1)
\end{IEEEeqnarray}     
where $ F_{\mathrm{v}_j^{i}}^{-1}$ is given by (\ref{eq_inverse_cdf}) and $ y^{+i}_j$ is $j-$th element of  $ y^{+i} \in \mathbb{R}^p$.    
 \RETURN $\mathbf{Y}^+ = \left \{ y^{+i} \in \mathbb{R}^p \; | \; i \in \{1,\cdots,N \} \right \}$. 
\end{algorithmic}
\end{algorithm} 
For a given value of $(\epsilon,\delta,d)$, Algorithm~\ref{algorithm_differential_private_approximation} is stated for a differentially private approximation of a data samples.   
\subsection{Differentially Private Semi-Supervised Transfer and Multi-Task Learning}\label{sec_4_2}
We consider a scenario of knowledge transfer from a dataset consisting of labeled samples from a domain (referred to as source domain) to another dataset consisting of mostly unlabelled samples and only a few labelled samples from another domain (referred to as target domain) such that both source and target datasets have been sampled from the same set of classes but in their respective domains. The aim is to transfer the knowledge extracted by a classifier trained using source dataset to the classifier of target domain such that privacy of source dataset is preserved. Let $\{\mathbf{Y}^{sr}_c \}_{c=1}^C$ be the labelled source dataset where $\mathbf{Y}^{sr}_c = \{ y^{i,c}_{sr} \in \mathbb{R}^{p_{sr}} \; | \; i \in \{1,\cdots,N^{sr}_c \} \}$ represents $c-$th labelled samples. The target dataset consist of a few labelled samples $\{\mathbf{Y}^{tg}_c \}_{c=1}^C$ (with $\mathbf{Y}^{tg}_c = \{ y^{i,c}_{tg} \in \mathbb{R}^{p_{tg}} \; | \; i \in \{1,\cdots,N^{tg}_c \} \}$) and another set of unlabelled samples $\mathbf{Y}^{tg}_* = \{ y^{i,*}_{tg} \in \mathbb{R}^{p_{tg}} \; | \; i \in \{1,\cdots,N^{tg}_* \} \}$. A generalized setting is considered where source and target data dimensions could be different, i.e., $p_{sr} \neq p_{tg}$. Our approach to semi-supervised transfer and multi-task learning consists of following steps:
\paragraph{Differentially private source domain classifier:} 
Since the noise adding mechanism (i.e. Result~\ref{result_optimal_noise_epsilon_delta_privacy}) is independent of the choice of algorithm operating on training data matrix, therefore any algorithm operating on noise added data samples will remain $(\epsilon,\delta)-$differentially private. That is, differential privacy remains invariant to any post-processing of noise added data samples. This allows us to build a differentially private classifier as stated in Algorithm~\ref{algorithm_private_classification}.
\begin{algorithm}
\caption{Variational learning of a differentially private classifier}
\label{algorithm_private_classification}
\begin{algorithmic}[1]
\REQUIRE Differentially private approximated dataset: $\mathbf{Y}^+ = \left \{ \mathbf{Y}_c^+\; | \;  c \in \{1,\cdots,C \} \right \}$; the subspace dimension $n \in \{1,\cdots,p\}$; ratio $r_{max} \in (0,1]$; the number of layers $L \in \mathbb{Z}_{+}$.  
\STATE Build a classifier, $ \{ \mathcal{P}_c^+  \}_{c=1}^C$, by applying Algorithm~\ref{algorithm_classification} on $\mathbf{Y}^+$ for the given $n$, $r_{max}$, and $L$.  
\RETURN $ \{ \mathcal{P}_c^+  \}_{c=1}^C$. 
\end{algorithmic}
\end{algorithm} 

For a given differential privacy parameters: $d,\epsilon,\delta$; Algorithm~\ref{algorithm_differential_private_approximation} is applied on $\mathbf{Y}^{sr}_c$ to obtain the differentially private approximated data samples, $\mathbf{Y}^{+sr}_c = \{ y^{+i,c}_{sr} \in \mathbb{R}^{p_{sr}} \; | \; i \in \{1,\cdots,N^{sr}_c \}\}$, for all $c \in \{1,\cdots,C \}$. Algorithm~\ref{algorithm_private_classification} is applied on $\{\mathbf{Y}^{+sr}_c \}_{c=1}^C$ to build a differentially private source domain classifier characterized by parameters sets $ \{ \mathcal{P}_c^{+sr}  \}_{c=1}^C$.     
\paragraph{Differentially private source domain latent subspace transformation-matrix} 
For a lower-dimensional representation of both source and target samples, a subspace dimension, $n_{st}  \in \{1,2,\cdots,\min(p_{sr},p_{tg})\}$, is chosen. Let  $V^{+sr} \in \mathbb{R}^{n_{st}  \times p_{sr}}$ be the transformation-matrix with its $i-$th row equal to transpose of eigenvector corresponding to $i-$th largest eigenvalue of sample covariance matrix computed on source samples. 
\paragraph{Target domain latent subspace transformation-matrix}
Let $V^{tg} \in \mathbb{R}^{n_{st} \times p_{tg}}$ be the transformation-matrix with its $i-$th row equal to transpose of eigenvector corresponding to $i-$th largest eigenvalue of sample covariance matrix computed on target samples. 
\paragraph{Subspace alignment for heterogenous domains}
For the case of heterogenous source and target domains (i.e. $p_{sr} \neq p_{tg}$), we follow \emph{subspace alignment} approach where a target sample is first aligned to source data in subspace followed by a linear transformation to source-data-space. A target sample can be mapped to source-data-space via following transformation: 
\begin{IEEEeqnarray}{CCl}
 y_{tg \rightarrow sr} & : = & \left \{\begin{array}{ll}  y_{tg}, & p_{sr} = p_{tg} \\
(V^{+sr})^TV^{tg}y_{tg}, & p_{sr} \neq p_{tg}
\end{array} \right. 
\end{IEEEeqnarray}   
Both labelled and unlabelled target datasets are transformed to define the following sets: 
\begin{IEEEeqnarray}{CCl}
\label{eq_738522.553647}\mathbf{Y}^{tg \rightarrow sr}_c & := & \{ y_{tg \rightarrow sr}   \; | \; y_{tg} \in \mathbf{Y}^{tg}_c \} \\
\label{eq_738522.553880} \mathbf{Y}^{tg \rightarrow sr}_* & := & \{ y_{tg \rightarrow sr} \; | \; y_{tg} \in \mathbf{Y}^{tg}_* \}.
\end{IEEEeqnarray}    
\paragraph{Building of target domain classifier} 
Our idea is to iteratively build target domain classifier for predicting the labels of unlabelled target data samples. The $k-$th iteration, for $k \in \{1,\cdots,it\_{max} \}$, consists of following updates:  
  \begin{IEEEeqnarray}{CCl}
    \label{eq_iterative_target_classification}  \{ \mathcal{P}_c^{tg}|_k  \}_{c=1}^C
 & = &   \text{Algorithm~\ref{algorithm_classification}}\left( \left \{\mathbf{Y}^{tg \rightarrow sr}_{c} \cup \mathbf{Y}^{tg \rightarrow sr}_{*,c}|_{k-1} \right \}_{c=1}^C, n|_k, r_{max},L \right) \\
  \label{eq_iterative_target_samples_class_c} \mathbf{Y}^{tg \rightarrow sr}_{*,c}|_{k} & = & \left \{ y^{i,*}_{tg \rightarrow sr} \in \mathbf{Y}^{tg \rightarrow sr}_*  \; | \; \mathcal{C}(y^{i,*}_{tg \rightarrow sr};\{ \mathcal{P}_c^{tg}|_{k} \}_{c=1}^C)  = c,\;  i \in \{1,\cdots,N^{tg}_* \} \right \}
  \end{IEEEeqnarray}  
where $\left\{n|_1,n|_2,\cdots \right \}$ is a monotonically non-decreasing sequence. The reason for $n$ to follow a monotonically non-decreasing curve during the iterations is following:
\begin{itemize}
\item[] {\it We intend to use higher-level data features during initial iterations for updating the predicted-labels of unlabeled target data samples and as the number of iterations increases more and more lower-level data features are intended to be included in the process of updating the predicted-labels. Since the lower values of $n$ lead to modeling of higher-level data features and higher values lead to modeling of lower-level data features (as illustrated in Fig.~\ref{fig_deep_learning_3}), $n$ values are chosen as to form a monotonically non-decreasing sequence.}        
\end{itemize}   
\paragraph{source2target model}
The target samples associated to a class can be filtered through the source domain autoencoder associated to the same class for defining the following dataset:
   \begin{IEEEeqnarray}{CCl}
\label{eq_738522.588685}\mathcal{D}   &:= & \left \{ \left( \widehat{\mathcal{WD}}(y;\mathcal{P}_c^{+sr}), y \right)  \; | \; y \in \left \{ \mathbf{Y}^{tg \rightarrow sr}_{c} \cup \mathbf{Y}^{tg \rightarrow sr}_{*,c}|_{it\_{max}} \right \},\; c \in \left \{1,\cdots,C  \right \} \right \}
  \end{IEEEeqnarray}   
where $\widehat{\mathcal{WD}}(\cdot;\cdot)$ is defined as in~(\ref{eq_738500.4495}), $\mathbf{Y}^{tg \rightarrow sr}_{c}$ is defined as in~(\ref{eq_738522.553647}), and $\mathbf{Y}^{tg \rightarrow sr}_{*,c}$ is defined as in (\ref{eq_iterative_target_samples_class_c}). Here, $\widehat{\mathcal{WD}}(y;\mathcal{P}_c^{+sr})$, where $y \in \{ \mathbf{Y}^{tg \rightarrow sr}_{c} \cup \mathbf{Y}^{tg \rightarrow sr}_{*,c}|_{it\_{max}} \}$, is a representation of a $c-$th labelled traget sample $y$ in the source domain $c-$th labelled data space represented by the wide CDMMA $\mathcal{P}_c^{+sr}$. The mapping from source to target domain can be learned via building a variational membership-mappings based model using Algorithm~\ref{algorithm_basic_learning} on the dataset $\mathcal{D} $. That is,
    \begin{IEEEeqnarray}{CCl}
 \label{eq_738522.588899} \mathbb{M}^{sr\rightarrow tg}& = &    \text{Algorithm~\ref{algorithm_basic_learning}}\left(\mathcal{D}, M_{max} \right) \\
 M_{max} & = & \min(\lceil N^{tg}/2 \rceil,1000) 
   \end{IEEEeqnarray} 
where $N^{tg} = | \mathcal{D} |$ is the total number of target samples.     
\paragraph{A transfer and multi-task learning scenario:}
Both source and target domain classifiers are combined with source2target model for predicting the label associated to a target sample $y_{tg \rightarrow sr}$ as 
  \begin{IEEEeqnarray}{CCl}
\nonumber  \hat{c}(y_{tg \rightarrow sr};\{ \mathcal{P}_c^{tg} \}_{c=1}^C, \{ \mathcal{P}_c^{+sr} \}_{c=1}^C,  \mathbb{M}^{sr\rightarrow tg})  & = & \arg\;\min_{c \:{\in}\: \{1,2,\cdots,C \}}\;  \left \{ \min\left(\left \| y_{tg \rightarrow sr} - \widehat{\mathcal{WD}}(y_{tg \rightarrow sr};\mathcal{P}_c^{tg})  \right \|^2, \right. \right. \\
\nonumber  && \left. \left. \left \| y_{tg \rightarrow sr}  - \hat{y} \left( \widehat{\mathcal{WD}}(y_{tg \rightarrow sr};\mathcal{P}_c^{+sr}) ; \mathbb{M}^{sr\rightarrow tg} \right) \right \|^2, \right. \right. \\
\label{eq_predicted_target_label_multitask} && \left. \left. \left \| y_{tg \rightarrow sr} - \widehat{\mathcal{WD}}(y_{tg \rightarrow sr};\mathcal{P}_c^{+sr})  \right \|^2   \right) \right \}. \IEEEeqnarraynumspace
\end{IEEEeqnarray}
where $\hat{y} \left( \cdot ; \mathbb{M}^{sr\rightarrow tg} \right)$ is the output of source2target model computed using (\ref{eq_738124.770095}). That is, $y_{tg \rightarrow sr}$ is assigned the $c-$th class label, if
\begin{itemize}
\item the autoencoder associated to $c-$th class of target data space (which is characterized by set of parameters $\mathcal{P}_c^{tg}$) could best reconstruct $y_{tg \rightarrow sr}$, or
\item the output of the source2target model with the input as representation of $y_{tg \rightarrow sr}$ in source domain $c-$th labelled data space could best reconstruct $y_{tg \rightarrow sr}$, or
\item the differentially private autoencoder associated to $c-$th class of source data space (which is characterized by set of parameters $\mathcal{P}_c^{+sr}$) could best reconstruct $y_{tg \rightarrow sr}$.   
\end{itemize}      
\section{Experiments}\label{sec_5}
Differentially private transferrable learning methodology was implemented using MATLAB R2017b. The experiments have been made on an iMac (M1, 2021) machine with 8 GB RAM. The implementational details for the method are described as below:    
\begin{itemize}
\item  We study experimentally the differential privacy (Definition~\ref{def_differential_privacy}) of the source domain training data such that for all $1-$adjacent training data matrices, the absolute value of privacy-loss incurred by observing the output of any computation algorithm will be bounded by $\epsilon$ with probability at least $1-\delta$. That is, $d$ is taken as equal to $1$ for defining adjacent matrices in Definition~\ref{def_differential_privacy}.
\item Algorithm~\ref{algorithm_differential_private_approximation}, for a given $d$ and $(\epsilon,\delta)$, is applied to obtain a differentially private approximation of source dataset.  
\item Differentially private source domain classifier is built using Algorithm~\ref{algorithm_private_classification} taking subspace dimension as equal to $\min(20,p_{sr})$ (where $p_{sr}$ is the dimension of source data samples), ratio $r_{max}$ as equal to 0.5, and number of layers as equal to 5. 
\item Differentially private source domain latent subspace transformation-matrix is computed with $n_{st} = \min(\lceil p_{sr}/2 \rceil,p_{tg})$, where $p_{tg}$ is the dimension of target data samples.
\item Initial target domain classifier is built using Algorithm~\ref{algorithm_classification} on labelled target samples taking subspace dimension as equal to $\min(20,\min_{1\leq c \leq C}\{ N^{tg}_c\}-1)$ (where $N^{tg}_c$ is the number of $c-$th class labelled target samples), ratio $r_{max}$ as equal to 1, and number of layers as equal to 1. 
\item The target domain classifier is updated using (\ref{eq_iterative_target_classification}) and (\ref{eq_iterative_target_samples_class_c}) till 4 iterations taking the monotonically non-decreasing subspace dimension $n$ sequence as $\{5,10,15,20\}$ and $r_{max = 0.5}$. 
\item The label associated to a target data point is predicted under transfer and multi-task learning scenarios using (\ref{eq_predicted_target_label_multitask}). 
\end{itemize}
\subsection{Demonstrative Examples Using MNIST and USPS Datasets}\label{sec_5_1}
\subsubsection{MNIST dataset}
\begin{figure}[bt]
\centering
\includegraphics[width = \textwidth]{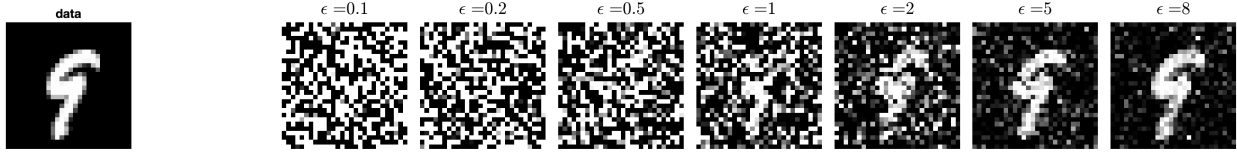}
\caption{An example of the noise added to a randomly selected sample from MNIST dataset for different values of $\epsilon$ with $\delta = 1\mathrm{e}{-5}$.}
\label{fig_examples_noise}
\end{figure}

Our first experiment is on the widely used MNIST digits dataset containing $28 \times 28$ sized images divided into training set of 60000 images and testing set of 10000 images. The images' pixel values were divided by 255 to normalize the values in the range from $0$ to $1$. The $28 \times 28$ normalized values of each image were flattened to an equivalent $784-$dimensional data vector. The transfer learning experiment was carried in the same setting as in~\cite{conf/iclr/PapernotAEGT17} where 60000 training samples constituted the source dataset; a set of 9000 test samples constituted target dataset, and the performance was evaluated on the remaining 1000 test samples. Out of 9000 target samples, only 10 samples per class were labelled and rest 8900 target samples remained as unlabelled. 
\paragraph{What is the sufficiently low value of privacy-loss bound?}
A lower privacy-loss bound implies a larger amount of noise being added to data samples. For an interpretation of the privacy-loss bound $\epsilon$ in terms of amount of noise required to be added to preserve data's privacy, the examples of noise added samples corresponding to different values of $\epsilon$ are provided in Fig.~\ref{fig_examples_noise}.
It is observed from Fig.~\ref{fig_examples_noise} that $\epsilon = 0.1$ is sufficiently low to preserve privacy in this case. Thus, the experiments were carried out with privacy-loss bound $\epsilon = 0.1$ while keeping failure probability fixed at $\delta = 1\mathrm{e}{-5}$. 
\paragraph{Competitive performance}
Table~\ref{table_results_MNIST} reports the experimental results. The proposed method is able to learn $95.1\%$ accurate model together with providing $(0.1,1\mathrm{e}{-5})-$differential privacy guarantee, which is a better result than the existing result~\cite{10.1145/2976749.2978318} of achieving $90\%$ accuracy for $(0.5,1\mathrm{e}{-5})-$differential privacy on MNIST dataset.  
\begin{table}
\centering
 \caption{Privacy and utility results on MNIST dataset. The second column reports the privacy-loss bound $\epsilon$ and failure probability $\delta$ of $(\epsilon,\delta)-$differential privacy guarantee.}
 \label{table_results_MNIST}
  {%
    \begin{tabular}{ccc}  
    \hline 
    \bfseries Method & \bfseries $(\epsilon,\delta)$ & \bfseries classification accuracy \\  
    \hline 
        proposed transfer and multi-task learning & $(0.1,1\mathrm{e}{-5})$  & 95.1\% \\
                          \cite{10.1145/2976749.2978318} & $(0.5,1\mathrm{e}{-5})$ & 90\% \\ 
        \hline 
    \end{tabular}  
  }  
\end{table}           
\subsubsection{Learning across heterogeneous MNIST and USPS domains}
We considered a problem of heterogeneous transfer learning between MNIST to USPS dataset. The USPS is another dataset that has 7291 training and 2007 test images of digits where each image has $16 \times16$ (=256) grayscale pixels. 
\paragraph{MNIST$\rightarrow$USPS}
The aim of this experiment was to study how privacy-preservation affect transferring knowledge from a higher resolution and more varied MNIST dataset to USPS dataset. The MNIST$\rightarrow$USPS semi-supervised transfer learning problem was previously studied in~\cite{Belhaj2018DeepVT}. For a comparison, MNIST$\rightarrow$USPS problem was considered in the same experimental setting as in~\cite{Belhaj2018DeepVT} where only 100 target samples were labelled and remaining 7191 samples remained as unlabelled. The experiments were carried out at privacy-loss bound $\epsilon \in \{0.1, 1\}$ while keeping failure probability fixed at $\delta = 1\mathrm{e}{-5}$. Further, the non-private version of the proposed method corresponding to the case of $\epsilon = \infty$ was also considered. The performance was evaluated on target domain testing dataset in-terms of classification accuracy. 
\begin{table}
\centering
 \caption{Results of 10 independent MNIST$\rightarrow$USPS experiments expressed in average accuracy $\pm$ standard deviation.}
 \label{table_results_MNIST_USPS_2}
  {%
    \begin{tabular}{cc}  
    \hline 
  \bfseries method & \bfseries accuracy (in \%) \\  
    \hline 
   $(0.1,1\mathrm{e}{-5})-$differentially private proposed & $92.23 \pm 0.87$ \\
      $(1,1\mathrm{e}{-5})-$differentially private proposed & $92.28 \pm 0.62$  \\ 
      non private proposed & $92.37 \pm 0.70$ \\
non private Deep Variational Transfer~\cite{Belhaj2018DeepVT} &  $92.03 \pm 0.38$ \\
        \hline 
    \end{tabular}  
  }  
\end{table}     

Table~\ref{table_results_MNIST_USPS_2} reports the results of 10 independent MNIST$\rightarrow$USPS experiments. As observed in Table~\ref{table_results_MNIST_USPS_2}, the proposed method, despite being privacy-preserving and having not required an access to source data samples, performs comparable to the Deep Variational Transfer (a variational autoencoder that transfers knowledge across domains using a shared latent Gaussian mixture model) proposed in~\cite{Belhaj2018DeepVT}. Further, the proposed method's consistent performance over a wide range of privacy-loss bound $\epsilon$ (from 0.1 to 1) verifies the robustness of the target model towards the perturbations in source data caused by the privacy requirements demanded by source data owner. 
\paragraph{Effect of labelled target sample size}
\begin{table}
\centering
\caption{Effect of labelled target sample size on performance in USPS$\rightarrow$MNIST problem.}
\label{table_effect_labelled_sample_size}
  {%
 \begin{tabular}{ccc}
 \hline
 \bfseries method & \bfseries $\begin{array}{c} \mbox{number of} \\ \mbox{ labelled target samples} \end{array}$ & \bfseries $\begin{array}{c} \mbox{classification accuracy} \\ \mbox{on target testing data} \end{array}$  \\
 \hline
   $(0.1,1\mathrm{e}{-5})-$differentially private proposed & 100 & 92.29\% \\ 
      $(0.1,1\mathrm{e}{-5})-$differentially private proposed & 200 & 95.86\% \\ 
            $(0.1,1\mathrm{e}{-5})-$differentially private proposed & 300 & 97.31\% \\ 
  $(0.1,1\mathrm{e}{-5})-$differentially private proposed & 400 & 97.63\% \\           
       $(0.1,1\mathrm{e}{-5})-$differentially private proposed & 500 & 97.99\% \\ \hline
 \end{tabular}   
  }
\end{table}
To study the effect of number of labelled target samples, USPS$\rightarrow$MNIST problem is considered with number of labelled target samples varying from $100$ to $500$. Table~\ref{table_effect_labelled_sample_size} reports the classification accuracy on target testing dataset as the number of labelled target samples is varied. It is verified that the proposed approach to combine source and target domain classifiers, as in (\ref{eq_predicted_target_label_multitask}), leads to an increasing performance with increasing labelled target sample size while preserving the privacy of source domain data. 

\subsection{Comparisons Using Office and Caltech256 Datasets}\label{sec_5_2}
``Office+Caltech256'' dataset has 10 common categories of both Office and Caltech256 datasets. This dataset has been widely used~\cite{Hoffman2013EfficientLO,Herath_2017_CVPR,8362683,Hoffman2014} for evaluating multi-class accuracy performance in a standard domain adaptation setting with a small number of labelled target samples. The dataset has fours domains: \emph{amazon}, \emph{webcam}, \emph{dslr}, and \emph{caltech256}. We follow the experimental setup of~\cite{Hoffman2013EfficientLO,Herath_2017_CVPR,8362683,Hoffman2014}:    
\begin{enumerate}
\item the number of training samples per class in the source domain is 20 for \emph{amazon} and is 8 for other three domains;
\item the number of labelled samples per class in the target domain is 3 for all the four domains;
\item 20 random train/test splits are created and the performance on target domain test samples is averaged over 20 experiments.  
\end{enumerate}
Following~\cite{Herath_2017_CVPR}, the deep-net VGG-FC6 features are extracted from the images and the proposed method is compared with
\begin{enumerate}
\item SVM-t: A base-line is created using a linear SVM classifier trained using only the labelled target samples without transfer learning.
\item ILS (1-NN)~\cite{Herath_2017_CVPR}: This method learns an Invariant Latent Space (ILS) to reduce the discrepancy between domains and uses Riemannian optimization techniques to match statistical properties between samples projected into the latent space from different domains. 
\item CDLS~\cite{7780918}: The Cross-Domain Landmark Selection (CDLS) method derives a domain-invariant feature subspace for heterogeneous domain adaptation. 
\item MMDT~\cite{Hoffman2014}: The Maximum Margin Domain Transform (MMDT) method adapts max-margin classifiers in a multi-class manner by learning a shared component of the domain shift as captured by the feature transformation. 
\item HFA~\cite{6587717}: The Heterogeneous Feature Augmentation (HFA) method learns common latent subspace and a classifier under max-margin framework.
\item OBTL~\cite{8362683}: The Optimal Bayesian Transfer Learning (OBTL) method employs Bayesian framework to transfer learning through modeling of a joint prior probability density function for feature-label distributions of the source and target domains.
\end{enumerate}
The ``Office+Caltech256'' dataset has been previously studied in \cite{Hoffman2013EfficientLO,Herath_2017_CVPR,8362683,Hoffman2014} using SURF features. Therefore, the state-of-art results on this dataset using SURF features are additionally considered for a comparison. There are in total 4 domains associated to ``Office+Caltech256'' dataset. Taking a domain as source and other domain as target, 12 different transfer learning experiments can be performed on these 4 domains. Table~\ref{table_amazon_2_caltech}, Table~\ref{table_amazon_2_dslr}, Table~\ref{table_amazon_2_webcam}, Table~\ref{table_caltech_2_amazon}, Table~\ref{table_caltech_2_dslr}, Table~\ref{table_caltech_2_webcam}, Table~\ref{table_dslr_2_amazon}, Table~\ref{table_dslr_2_caltech}, Table~\ref{table_dslr_2_webcam}, Table~\ref{table_webcam_2_amazon}, Table~\ref{table_webcam_2_caltech}, and Table~\ref{table_webcam_2_dslr} report the results and the first two best performances have been marked. 
\begin{table}
\centering
 \caption{Accuracy (in \%, averaged over 20 experiments) obtained in \emph{amazon}$\rightarrow$\emph{caltech256} semi-supervised transfer learning experiments.}
 \label{table_amazon_2_caltech}
  {%
    \begin{tabular}{ccc}  
    \hline 
  \bfseries method & \bfseries feature type &  \bfseries accuracy (\%)  \\  
    \hline 
 $(0.1,1\mathrm{e}{-5})-$differentially private proposed   & VGG-FC6  & \underline{80.6}   \\
 SVM-t (without knowledge transfer) & VGG-FC6 &  73.4  \\  
non-private ILS (1-NN) & VGG-FC6 &  \underline{\textbf{83.3}}  \\
non-private CDLS & VGG-FC6 &  78.1  \\
non-private MMDT & VGG-FC6 & 78.7  \\
non-private HFA & VGG-FC6 &  75.5  \\
non-private OBTL & SURF &  41.5  \\
non-private ILS (1-NN) & SURF &  43.6  \\
non-private CDLS & SURF &  35.3  \\
non-private MMDT & SURF &  36.4 \\
non-private HFA & SURF & 31.0  \\
        \hline 
    \end{tabular}  
  }  
\end{table}   
\begin{table}
\centering
 \caption{Accuracy (in \%, averaged over 20 experiments) obtained in \emph{amazon}$\rightarrow$\emph{dslr} semi-supervised transfer learning experiments.}
 \label{table_amazon_2_dslr}
  {%
    \begin{tabular}{ccc}  
    \hline 
  \bfseries method & \bfseries feature type &  \bfseries accuracy (\%)  \\  
    \hline 
  $(0.1,1\mathrm{e}{-5})-$differentially private proposed   & VGG-FC6 &    \underline{\textbf{91.2}}      \\
 SVM-t (without knowledge transfer)& VGG-FC6 & \underline{90.0} \\  
non-private ILS (1-NN) & VGG-FC6 &  87.7 \\
non-private CDLS & VGG-FC6 &  86.9   \\
non-private MMDT & VGG-FC6 &  77.1   \\
non-private HFA & VGG-FC6 &  87.1   \\
non-private OBTL & SURF &  60.2   \\
non-private ILS (1-NN) & SURF &  49.8   \\
non-private CDLS & SURF &  60.4   \\
non-private MMDT & SURF &  56.7   \\
non-private HFA & SURF &  55.1  \\
        \hline 
    \end{tabular}  
  }  
\end{table}   
\begin{table}
\centering
 \caption{Accuracy (in \%, averaged over 20 experiments) obtained in \emph{amazon}$\rightarrow$\emph{webcam} semi-supervised transfer learning experiments.}
 \label{table_amazon_2_webcam}
  {%
    \begin{tabular}{ccc}  
    \hline 
  \bfseries method & \bfseries feature type & \bfseries accuracy (\%)  \\  
    \hline 
 $(0.1,1\mathrm{e}{-5})-$differentially private proposed   & VGG-FC6 &     89.5     \\
 SVM-t (without knowledge transfer)& VGG-FC6 &  86.9  \\  
non-private ILS (1-NN) & VGG-FC6 &  \underline{90.7}   \\
non-private CDLS & VGG-FC6 &  \underline{\textbf{91.2}}  \\
non-private MMDT & VGG-FC6 &  82.5   \\
non-private HFA & VGG-FC6 &  87.9   \\
non-private OBTL & SURF &  72.4   \\
non-private ILS (1-NN) & SURF &  59.7   \\
non-private CDLS & SURF &  68.7   \\
non-private MMDT & SURF &   64.6  \\
non-private HFA & SURF &  57.4  \\
        \hline 
    \end{tabular}  
  }  
\end{table}   
\begin{table}
\centering
 \caption{Accuracy (in \%, averaged over 20 experiments) obtained in \emph{caltech256}$\rightarrow$\emph{amazon} semi-supervised transfer learning experiments.}
 \label{table_caltech_2_amazon}
  {%
    \begin{tabular}{ccc}  
    \hline 
  \bfseries method & \bfseries feature type &  \bfseries accuracy (\%)  \\  
    \hline 
 $(0.1,1\mathrm{e}{-5})-$differentially private proposed   & VGG-FC6 &    \underline{\textbf{91.5}}   \\
 SVM-t (without knowledge transfer) & VGG-FC6 &   84.4  \\  
non-private ILS (1-NN) & VGG-FC6 &  \underline{89.7}   \\
non-private CDLS & VGG-FC6 &  88.0  \\
non-private MMDT & VGG-FC6 &  85.9  \\
non-private HFA & VGG-FC6 &  86.2   \\
non-private OBTL & SURF &   54.8  \\
non-private ILS (1-NN) & SURF &   55.1   \\
non-private CDLS & SURF &   50.9  \\
non-private MMDT & SURF &  49.4   \\
non-private HFA & SURF &  43.8  \\
        \hline 
    \end{tabular}  
  }  
\end{table}   
\begin{table}
\centering
 \caption{Accuracy (in \%, averaged over 20 experiments) obtained in \emph{caltech256}$\rightarrow$\emph{dslr} semi-supervised transfer learning experiments.}
 \label{table_caltech_2_dslr}
  {%
    \begin{tabular}{ccc}  
    \hline 
  \bfseries method & \bfseries feature type &  \bfseries accuracy (\%)  \\  
    \hline 
 $(0.1,1\mathrm{e}{-5})-$differentially private proposed  & VGG-FC6 &    \underline{\textbf{91.6}} \\
 SVM-t (without knowledge transfer) & VGG-FC6 &    \underline{90.2} \\  
non-private ILS (1-NN) & VGG-FC6 &   86.9  \\
non-private CDLS & VGG-FC6 &  86.3  \\
non-private MMDT & VGG-FC6 &   77.9   \\
non-private HFA & VGG-FC6 &  87.0    \\
non-private OBTL & SURF &  61.5   \\
non-private ILS (1-NN) & SURF &   56.2  \\
non-private CDLS & SURF &   59.8  \\
non-private MMDT & SURF &   56.5  \\
non-private HFA & SURF &   55.6 \\
        \hline 
    \end{tabular}  
  }  
\end{table}   
\begin{table}
\centering
 \caption{Accuracy (in \%, averaged over 20 experiments) obtained in \emph{caltech256}$\rightarrow$\emph{webcam} semi-supervised transfer learning experiments.}
 \label{table_caltech_2_webcam}
  {%
    \begin{tabular}{ccc}  
    \hline 
  \bfseries method & \bfseries feature type &  \bfseries accuracy (\%)  \\  
    \hline 
 $(0.1,1\mathrm{e}{-5})-$differentially private proposed   & VGG-FC6 &    \underline{\textbf{91.6}}   \\
 SVM-t (without knowledge transfer) & VGG-FC6 &   88.8  \\  
non-private ILS (1-NN) & VGG-FC6 &  \underline{91.4}   \\
non-private CDLS & VGG-FC6 &  89.7   \\
non-private MMDT & VGG-FC6 &  82.8   \\
non-private HFA & VGG-FC6 &  86.0    \\
non-private OBTL & SURF &  71.1   \\
non-private ILS (1-NN) & SURF &   62.9  \\
non-private CDLS & SURF &  66.3   \\
non-private MMDT & SURF &   63.8  \\
non-private HFA & SURF &  58.1  \\
        \hline 
    \end{tabular}  
  }  
\end{table}   
\begin{table}
\centering
 \caption{Accuracy (in \%, averaged over 20 experiments) obtained in \emph{dslr}$\rightarrow$\emph{amazon} semi-supervised transfer learning experiments.}
 \label{table_dslr_2_amazon}
  {%
    \begin{tabular}{ccc}  
    \hline 
  \bfseries method & \bfseries feature type &  \bfseries accuracy (\%)   \\  
    \hline 
 $(0.1,1\mathrm{e}{-5})-$differentially private proposed   & VGG-FC6 &   \underline{\textbf{90.7}} \\
 SVM-t (without knowledge transfer) & VGG-FC6 &   84.4  \\  
non-private ILS (1-NN) & VGG-FC6 &  \underline{88.7}   \\
non-private CDLS & VGG-FC6 &   88.1  \\
non-private MMDT & VGG-FC6 &   83.6  \\
non-private HFA & VGG-FC6 &   85.9   \\
non-private OBTL & SURF &  54.4  \\
non-private ILS (1-NN) & SURF &   55.0  \\
non-private CDLS & SURF &  50.7  \\
non-private MMDT & SURF &  46.9  \\
non-private HFA & SURF &   42.9  \\
        \hline 
    \end{tabular}  
  }  
\end{table}   
 \begin{table}
\centering
 \caption{Accuracy (in \%, averaged over 20 experiments) obtained in \emph{dslr}$\rightarrow$\emph{caltech256} semi-supervised transfer learning experiments.}
 \label{table_dslr_2_caltech}
  {%
    \begin{tabular}{ccc}  
    \hline 
  \bfseries method & \bfseries feature type & \bfseries accuracy (\%)  \\  
    \hline 
 $(0.1,1\mathrm{e}{-5})-$differentially private proposed  & VGG-FC6 &    \underline{\textbf{81.4}}    \\
 SVM-t (without knowledge transfer) & VGG-FC6 &   73.6  \\  
non-private ILS (1-NN) & VGG-FC6 &   \underline{\textbf{81.4}}  \\
non-private CDLS & VGG-FC6 &  \underline{77.9}   \\
non-private MMDT & VGG-FC6 &  71.8  \\
non-private HFA & VGG-FC6 &  74.8   \\
non-private OBTL & SURF &  40.3  \\
non-private ILS (1-NN) & SURF &  41.0  \\
non-private CDLS & SURF &  34.9  \\
non-private MMDT & SURF &  34.1  \\
non-private HFA & SURF &   30.9  \\
        \hline 
    \end{tabular}  
  }  
\end{table}   
 \begin{table}
\centering
 \caption{Accuracy (in \%, averaged over 20 experiments) obtained in \emph{dslr}$\rightarrow$\emph{webcam} semi-supervised transfer learning experiments.}
 \label{table_dslr_2_webcam}
  {%
    \begin{tabular}{cccc}  
    \hline 
  \bfseries method & \bfseries feature type &  \bfseries accuracy (\%)  \\  
    \hline 
 $(0.1,1\mathrm{e}{-5})-$differentially private proposed  & VGG-FC6 &   88.7    \\
 SVM-t (without knowledge transfer) & VGG-FC6 &  86.9   \\  
non-private ILS (1-NN) & VGG-FC6 &   \underline{\textbf{95.5}}  \\
non-private CDLS & VGG-FC6 &  \underline{90.7}  \\
non-private MMDT & VGG-FC6 &   86.1 \\
non-private HFA & VGG-FC6 &   86.9  \\
non-private OBTL & SURF &  83.2 \\
non-private ILS (1-NN) & SURF &  80.1   \\
non-private CDLS & SURF &  68.5  \\
non-private MMDT & SURF &  74.1  \\
non-private HFA & SURF &  60.5   \\
        \hline 
    \end{tabular}  
  }  
\end{table}   
\begin{table}
\centering
 \caption{Accuracy (in \%, averaged over 20 experiments) obtained in \emph{webcam}$\rightarrow$\emph{amazon} semi-supervised transfer learning experiments.}
 \label{table_webcam_2_amazon}
  {%
    \begin{tabular}{cccc}  
    \hline 
  \bfseries method & \bfseries feature type &  \bfseries accuracy (\%)  \\  
    \hline 
 $(0.1,1\mathrm{e}{-5})-$differentially private proposed  & VGG-FC6 &   \underline{\textbf{92.0}}    \\
SVM-t (without knowledge transfer) & VGG-FC6 & 84.8    \\  
non-private ILS (1-NN) & VGG-FC6 &  \underline{88.8}   \\
non-private CDLS & VGG-FC6 &  87.4  \\
non-private MMDT & VGG-FC6 &  84.7  \\
non-private HFA & VGG-FC6 &  85.1  \\
non-private OBTL & SURF &   55.0 \\
non-private ILS (1-NN) & SURF & 54.3   \\
non-private CDLS & SURF &  51.8  \\
non-private MMDT & SURF &  47.7  \\
non-private HFA & SURF &  56.5   \\
        \hline 
    \end{tabular}  
  }  
\end{table}   
 \begin{table}
\centering
 \caption{Accuracy (in \%, averaged over 20 experiments) obtained in \emph{webcam}$\rightarrow$\emph{caltech256} semi-supervised transfer learning experiments.}
 \label{table_webcam_2_caltech}
  {%
    \begin{tabular}{cccc}  
    \hline 
  \bfseries method & \bfseries feature type &  \bfseries accuracy (\%)  \\  
    \hline 
 $(0.1,1\mathrm{e}{-5})-$differentially private proposed   & VGG-FC6 &   \underline{82.3}    \\
 SVM-t (without knowledge transfer) & VGG-FC6 & 75.0    \\  
non-private ILS (1-NN) & VGG-FC6 &  \underline{\textbf{82.8}}   \\
non-private CDLS & VGG-FC6 &  78.2  \\
non-private MMDT & VGG-FC6 &  73.6  \\
non-private HFA & VGG-FC6 &  74.4   \\
non-private OBTL & SURF &  37.4   \\
non-private ILS (1-NN) & SURF &   38.6   \\
non-private CDLS & SURF &  33.5   \\
non-private MMDT & SURF & 32.2   \\
non-private HFA & SURF &   29.0  \\
        \hline 
    \end{tabular}  
  }  
\end{table}
\begin{table}
\centering
 \caption{Accuracy (in \%, averaged over 20 experiments) obtained in \emph{webcam}$\rightarrow$\emph{dslr} semi-supervised transfer learning experiments.}
 \label{table_webcam_2_dslr}
  {%
    \begin{tabular}{ccc}  
    \hline 
  \bfseries method & \bfseries feature type &  \bfseries accuracy (\%)  \\  
    \hline 
 $(0.1,1\mathrm{e}{-5})-$differentially private proposed   & VGG-FC6 &    \underline{89.6}     \\
 SVM-t (without knowledge transfer) & VGG-FC6 &  88.9 \\  
non-private ILS (1-NN) & VGG-FC6 &  \underline{\textbf{94.5}}  \\
non-private CDLS & VGG-FC6 &  88.5   \\
non-private MMDT & VGG-FC6 &   85.1  \\
non-private HFA & VGG-FC6 &   87.3 \\
non-private OBTL & SURF &  75.0   \\
non-private ILS (1-NN) & SURF  & 70.8    \\
non-private CDLS & SURF &  60.7  \\
non-private MMDT & SURF &   67.0  \\
non-private HFA & SURF &   56.5  \\
        \hline 
    \end{tabular}  
  }  
\end{table}
\begin{table}
\centering
\caption{Comparison of the methods on ``Office+Caltech256'' dataset.}
 \label{table_overall_performance}
 {
   \begin{tabular}{ccc}  
     \hline 
  \bfseries method &  \bfseries $\begin{array}{c} \mbox{number of experiments} \\ \mbox{in which method} \\ \mbox{performed best} \end{array}$ & \bfseries $\begin{array}{c} \mbox{number of experiments} \\ \mbox{in which method} \\ \mbox{performed 2nd best} \end{array}$    \\  
    \hline 
  $(0.1,1\mathrm{e}{-5})-$differentially private proposed & 7 & 3 \\ 
  non-private ILS (1-NN) &  5 & 5 \\ 
  non-private CDLS & 1 & 2 \\
 \hline
   \end{tabular}
 }
\end{table}
\clearpage

Finally, Table~\ref{table_overall_performance} summarizes the overall performance of top three methods. As observed in Table~\ref{table_overall_performance}, the proposed method remains as best performing in maximum number of experiments. The most remarkable result observed is that the proposed transfer and multi-task learning method, despite ensuring privacy-loss bound to be as low as 0.1 and not requiring an access to source data samples, performs better than even the non-private methods.   

\section{Concluding Remarks}\label{sec_conclusion}
This study has outlined a novel approach to differentially private semi-supervised transfer and multi-task learning that exploits the variational deep mappings and an optimal noise adding mechanism for achieving a robustness of target model towards the perturbations in source data caused by the privacy requirements demanded by source data owner. A variational membership-mapping based approach was introduced that sufficiently addresses all of the requirements identified regarding the privacy-preserving transferrable deep learning problem. Numerous experiments were carried out using MNIST, USPS, Office, and Caltech256 datasets to verify the competitive performance of the proposed method. The experimental studies further verify that our approach is capable of achieving a low privacy-loss bound without letting the accuracy be much degraded.

\appendix

\section{Evaluation of Membership Function}\label{appendix_1}
Using (\ref{eq_pf_1001_student_t}), we have 
\begin{IEEEeqnarray*}{rCl}
\left< \log ( \mu_{\mathrm{y}_j;\mathrm{f}_j}(\tilde{\mathrm{y}}_j) ) \right>_{\mu_{\mathrm{f}_j;\mathrm{u}_j} } &  = & - 0.5 \beta \|\tilde{\mathrm{y}}_j- \bar{m}_{\mathrm{f}_j} \|^2 - 0.5\beta \frac{\nu + (\mathrm{u}_j)^T (K_{\mathrm{a}\mathrm{a}})^{-1} \mathrm{u}_j  - 2}{\nu + M - 2} Tr( \bar{K}_{\mathrm{x}\mathrm{x}})
\end{IEEEeqnarray*}
 where $Tr(\cdot)$ denotes the trace operator. Using (\ref{eq_pf_1002}) and (\ref{eq_pf_1003}),  
\begin{IEEEeqnarray}{rCl}
\nonumber  \left< \log ( \mu_{\mathrm{y}_j;\mathrm{f}_j}(\tilde{\mathrm{y}}_j))  \right>_{\mu_{\mathrm{f}_j;\mathrm{u}_j} }   &=&    - 0.5\beta \| \tilde{\mathrm{y}}_j  \|^2 + \beta (\tilde{\mathrm{y}}_j)^T K_{\mathrm{x}\mathrm{a}} (K_{\mathrm{a}\mathrm{a}})^{-1} \mathrm{u}_j  - 0.5\beta (\mathrm{u}_j)^T (K_{\mathrm{a}\mathrm{a}})^{-1} K_{\mathrm{x}\mathrm{a}}^T K_{\mathrm{x}\mathrm{a}} (K_{\mathrm{a}\mathrm{a}})^{-1} \mathrm{u}_j \\
\label{eq_FTNDL_1}  && - 0.5\beta \frac{\nu + (\mathrm{u}_j)^T (K_{\mathrm{a}\mathrm{a}})^{-1} \mathrm{u}_j  - 2}{\nu + M - 2} \left( Tr(K_{\mathrm{x}\mathrm{x}} )   - Tr((K_{\mathrm{a}\mathrm{a}})^{-1} K_{\mathrm{x}\mathrm{a}}^T K_{\mathrm{x}\mathrm{a}}) \right).
\end{IEEEeqnarray}
Using~(\ref{eq_738494.4689}),
\begin{IEEEeqnarray*}{rCl}
\mu_{\mathrm{y}_j;\mathrm{u}_j}(\tilde{\mathrm{y}}_j) & \propto &  \exp \left(  - 0.5\beta \| \tilde{\mathrm{y}}_j  \|^2 + \beta (\tilde{\mathrm{y}}_j)^T K_{\mathrm{x}\mathrm{a}} (K_{\mathrm{a}\mathrm{a}})^{-1} \mathrm{u}_j  - 0.5\beta (\mathrm{u}_j)^T (K_{\mathrm{a}\mathrm{a}})^{-1} K_{\mathrm{x}\mathrm{a}}^T K_{\mathrm{x}\mathrm{a}} (K_{\mathrm{a}\mathrm{a}})^{-1} \mathrm{u}_j \right. \\
&& \left. - 0.5\beta \frac{ (\mathrm{u}_j)^T (K_{\mathrm{a}\mathrm{a}})^{-1} \mathrm{u}_j }{\nu + M - 2} \left( Tr(K_{\mathrm{x}\mathrm{x}} )   - Tr((K_{\mathrm{a}\mathrm{a}})^{-1} K_{\mathrm{x}\mathrm{a}}^T K_{\mathrm{x}\mathrm{a}}) \right)  + \{/ (\tilde{\mathrm{y}}_j,\mathrm{u}_j)\}  \right)
\end{IEEEeqnarray*} 
where $ \{/ (\tilde{\mathrm{y}}_j,\mathrm{u}_j)\}$ represents all those terms which are independent of both $\tilde{\mathrm{y}}_j$ and $\mathrm{u}_j$. Define  
 \begin{IEEEeqnarray}{rCl}
\label{eq_hat_K_u_1000}\hat{K}_{\mathrm{u}_j} & = & \left((K_{\mathrm{a}\mathrm{a}})^{-1}  + \beta (K_{\mathrm{a}\mathrm{a}})^{-1}  K_{\mathrm{x}\mathrm{a}}^T K_{\mathrm{x}\mathrm{a}}  (K_{\mathrm{a}\mathrm{a}})^{-1}  + \beta \frac{Tr(K_{\mathrm{x}\mathrm{x}} )   - Tr((K_{\mathrm{a}\mathrm{a}})^{-1} K_{\mathrm{x}\mathrm{a}}^T K_{\mathrm{x}\mathrm{a}})}{\nu + M - 2} (K_{\mathrm{a}\mathrm{a}})^{-1}  \right)^{-1} \IEEEeqnarraynumspace \\
\label{eq_hat_m_u_1000} \hat{m}_{\mathrm{u}_j}(\tilde{\mathrm{y}}_j) & = & \beta \hat{K}_{\mathrm{u}_j} (K_{\mathrm{a}\mathrm{a}})^{-1} (K_{\mathrm{x}\mathrm{a}})^T \tilde{\mathrm{y}}_j  
\end{IEEEeqnarray}
to express $\mu_{\mathrm{y}_j;\mathrm{u}_j}(\tilde{\mathrm{y}}_j) $ as (\ref{eq_log_membership2}).

\section{Solution of Optimization Problem}\label{appendix_2}
A new objective functional is defined after excluding $\mathrm{u}_j-$independent terms and taking into account the integral constraint through a Lagrange multiplier $\gamma$:  
\begin{IEEEeqnarray}{rCl}
\nonumber  \mathcal{J} & = & \left<  (\mathrm{u}_j)^T \hat{K}_{\mathrm{u}_j}^{-1} \hat{m}_{\mathrm{u}_j }(\mathrm{y}_j) - 0.5 (\mathrm{u}_j)^T \hat{K}_{\mathrm{u}_j}^{-1} \mathrm{u}_j  + 0.5 (\mathrm{u}_j)^T  (K_{\mathrm{a}\mathrm{a}})^{-1}  \mathrm{u}_j   - \log ( \mu_{\mathrm{u}_j}(\mathrm{u}_j) ) -0.5(\mathrm{u}_j)^T (K_{\mathrm{a}\mathrm{a}})^{-1} \mathrm{u}_j  \right >_{\mu_{\mathrm{u}_j}} \\
&&   + \gamma \left\{ \int_{\mathbb{R}^M} \mu_{\mathrm{u}_j}(\mathrm{u}_j)   \, \dd \lambda^M(\mathrm{u}_j)  - C_{\mathrm{u}_j} \right \} \\
 \nonumber & = &  \frac{1}{C_{\mathrm{u}_j}} \int_{\mathbb{R}^M} \dd \lambda^M(\mathrm{u}_j)  \: \mu_{\mathrm{u}_j}(\mathrm{u}_j)  \left\{ (\mathrm{u}_j)^T \hat{K}_{\mathrm{u}_j}^{-1} \hat{m}_{\mathrm{u}_j }(\mathrm{y}_j) - 0.5 (\mathrm{u}_j)^T \hat{K}_{\mathrm{u}_j}^{-1} \mathrm{u}_j        - \log (\mu_{\mathrm{u}_j}(\mathrm{u}_j) ) \right\} \\
 && + \gamma \left\{ \int_{\mathbb{R}^M} \mu_{\mathrm{u}_j}(\mathrm{u}_j)   \, \dd \lambda^M(\mathrm{u}_j)   - C_{\mathrm{u}_j} \right \}
\end{IEEEeqnarray}
Setting the functional derivative of $\mathcal{J}$ w.r.t. $\mu_{\mathrm{u}_j}$ equal to zero, 
\begin{IEEEeqnarray}{rCl}
0 & = & \gamma + (1/C_{\mathrm{u}_j}) \left\{-1 - 0.5 (\mathrm{u}_j)^T \hat{K}_{\mathrm{u}_j}^{-1} \mathrm{u}_j + (\mathrm{u}_j)^T \hat{K}_{\mathrm{u}_j}^{-1} \hat{m}_{\mathrm{u}_j }(\mathrm{y}_j)   - \log(\mu_{\mathrm{u}_j}(\mathrm{u}_j) )    \right\}.
\end{IEEEeqnarray}
That is,
\begin{IEEEeqnarray}{rCl}
\mu_{\mathrm{u}_j}(\mathrm{u}_j)  & = & \exp(\gamma C_{\mathrm{u}_j} - 1) \exp \left( - 0.5 (\mathrm{u}_j)^T \hat{K}_{\mathrm{u}_j}^{-1} \mathrm{u}_j  + (\mathrm{u}_j)^T \hat{K}_{\mathrm{u}_j}^{-1} \hat{m}_{\mathrm{u}_j }(\mathrm{y}_j) \right).
\end{IEEEeqnarray}
The optimal value of $\gamma$ is obtained by solving $ \int_{\mathbb{R}^M} \mu_{\mathrm{u}_j}  \, \dd \lambda^M =C_{\mathrm{u}_j}$. This leads to
\begin{IEEEeqnarray}{rCl}
 \exp(\gamma C_{\mathrm{u}_j} - 1) \sqrt{ (2 \pi)^M / | \hat{K}_{\mathrm{u}_j}^{-1}  | } \exp\left( 0.5 \left(\hat{m}_{\mathrm{u}_j }(\mathrm{y}_j)\right)^T \hat{K}_{\mathrm{u}_j}^{-1} \hat{m}_{\mathrm{u}_j }(\mathrm{y}_j)  \right) & = & C_{\mathrm{u}_j}.
\end{IEEEeqnarray}
Thus, the optimal expression for $\mu_{\mathrm{u}_j}$ is given as
\begin{IEEEeqnarray}{rCl}
\mu_{\mathrm{u}_j}^*(\mathrm{u}_j) & = & C_{\mathrm{u}_j} \sqrt{  | \hat{K}_{\mathrm{u}_j}^{-1}  | / (2 \pi)^M} \exp\left(- 0.5 (\mathrm{u}_j - \hat{m}_{\mathrm{u}_j}(\mathrm{y}_j))^T \hat{K}_{\mathrm{u}_j}^{-1}(\mathrm{u}_j - \hat{m}_{\mathrm{u}_j}(\mathrm{y}_j)) \right).
\end{IEEEeqnarray}
Finally, $C_{\mathrm{u}_j}$ is chosen such that $\max_{\mathrm{u}_j}\mu_{\mathrm{u}_j}^*(\mathrm{u}_j) = 1$. This results in
\begin{IEEEeqnarray}{rCl}
\mu_{\mathrm{u}_j}^*(\mathrm{u}_j) & = & \exp\left(- 0.5 (\mathrm{u}_j - \hat{m}_{\mathrm{u}_j}(\mathrm{y}_j))^T \hat{K}_{\mathrm{u}_j}^{-1}(\mathrm{u}_j - \hat{m}_{\mathrm{u}_j}(\mathrm{y}_j)) \right).
\end{IEEEeqnarray}
Thus, $\left< \mathrm{u}_j \right>_{\mu_{\mathrm{u}_j}^*} = \hat{m}_{\mathrm{u}_j}(\mathrm{y}_j)$, and using (\ref{eq_hat_m_u_1000}), we get
\begin{IEEEeqnarray}{rCl}
 \label{eq_satguru_7_5}  \left< \mathrm{u}_j \right>_{\mu_{\mathrm{u}_j}^* } & = & \beta \hat{K}_{\mathrm{u}_j} (K_{\mathrm{a}\mathrm{a}})^{-1} (K_{\mathrm{x}\mathrm{a}})^T \mathrm{y}_j.
\end{IEEEeqnarray}
It follows from (\ref{eq_hat_K_u_1000}) and (\ref{eq_hat_m_u_1000}) that
\begin{IEEEeqnarray}{rCl}
\label{eq_738495.3777} \hat{K}_{\mathrm{u}_j}^{-1} - (K_{\mathrm{a}\mathrm{a}})^{-1}  & = & \beta (K_{\mathrm{a}\mathrm{a}})^{-1}  K_{\mathrm{x}\mathrm{a}}^T K_{\mathrm{x}\mathrm{a}} (K_{\mathrm{a}\mathrm{a}})^{-1}  + \beta \frac{Tr(K_{\mathrm{x}\mathrm{x}}) - Tr((K_{\mathrm{a}\mathrm{a}})^{-1} K_{\mathrm{x}\mathrm{a}}^TK_{\mathrm{x}\mathrm{a}})}{\nu+ M - 2} (K_{\mathrm{a}\mathrm{a}})^{-1}\\
 \label{eq_738495.3779} \hat{K}_{\mathrm{u}_j}^{-1}  \hat{m}_{\mathrm{u}_j}(\mathrm{y}_j) & = & \beta (K_{\mathrm{a}\mathrm{a}})^{-1} (K_{\mathrm{x}\mathrm{a}})^T \mathrm{y}_j.
\end{IEEEeqnarray}
Using (\ref{eq_738495.3777}) and (\ref{eq_738495.3779}) in (\ref{eq_scch_1}), we have
\begin{IEEEeqnarray}{rCl}
\nonumber \log( \mu_{\mathrm{y}_j;\mathrm{u}_j}(\tilde{\mathrm{y}}_j) )  &= & - 0.5\beta \| \tilde{\mathrm{y}}_j\|^2 + \beta (\mathrm{u}_j)^T (K_{\mathrm{a}\mathrm{a}})^{-1} (K_{\mathrm{x}\mathrm{a}})^T \tilde{\mathrm{y}}_j\\
\nonumber && - 0.5\beta(\mathrm{u}_j)^T \left\{ (K_{\mathrm{a}\mathrm{a}})^{-1}  K_{\mathrm{x}\mathrm{a}}^T K_{\mathrm{x}\mathrm{a}} (K_{\mathrm{a}\mathrm{a}})^{-1}  + \frac{Tr(K_{\mathrm{x}\mathrm{x}}) - Tr((K_{\mathrm{a}\mathrm{a}})^{-1} K_{\mathrm{x}\mathrm{a}}^TK_{\mathrm{x}\mathrm{a}})}{\nu+ M - 2} (K_{\mathrm{a}\mathrm{a}})^{-1} \right\} \mathrm{u}_j .
\end{IEEEeqnarray} 
Thus, $\left<  \log( \mu_{\mathrm{y}_j;\mathrm{u}_j}(\tilde{\mathrm{y}}_j) ) \right>_{\mu_{\mathrm{u}_j}^*} $ is given as 
\begin{IEEEeqnarray}{rCl}
\nonumber \lefteqn{\left<  \log( \mu_{\mathrm{y}_j;\mathrm{u}_j}(\tilde{\mathrm{y}}_j) )\right>_{\mu^*_{\mathrm{u}_j}}} \\
\nonumber & = & - 0.5\beta \| \tilde{\mathrm{y}}_j\|^2  + \beta\left(\hat{m}_{\mathrm{u}_j }(\mathrm{y}_j)\right)^T (K_{\mathrm{a}\mathrm{a}})^{-1} (K_{\mathrm{x}\mathrm{a}})^T \tilde{\mathrm{y}}_j \\
\nonumber &&  - 0.5\beta\left(\hat{m}_{\mathrm{u}_j }(\mathrm{y}_j)\right)^T \left\{ (K_{\mathrm{a}\mathrm{a}})^{-1}  K_{\mathrm{x}\mathrm{a}}^T K_{\mathrm{x}\mathrm{a}}   (K_{\mathrm{a}\mathrm{a}})^{-1}   + \frac{Tr(K_{\mathrm{x}\mathrm{x}}) - Tr((K_{\mathrm{a}\mathrm{a}})^{-1}  K_{\mathrm{x}\mathrm{a}}^T K_{\mathrm{x}\mathrm{a}} )}{\nu+ M - 2} (K_{\mathrm{a}\mathrm{a}})^{-1} \right\} \hat{m}_{\mathrm{u}_j }(\mathrm{y}_j) \\
\label{eq_log_membership3} && - 0.5\beta Tr\left( (K_{\mathrm{a}\mathrm{a}})^{-1}  K_{\mathrm{x}\mathrm{a}}^T K_{\mathrm{x}\mathrm{a}}  (K_{\mathrm{a}\mathrm{a}})^{-1} \hat{K}_{\mathrm{u}_j}   + \frac{Tr(K_{\mathrm{x}\mathrm{x}}) - Tr((K_{\mathrm{a}\mathrm{a}})^{-1}  K_{\mathrm{x}\mathrm{a}}^T K_{\mathrm{x}\mathrm{a}} )}{\nu+ M - 2} (K_{\mathrm{a}\mathrm{a}})^{-1} \hat{K}_{\mathrm{u}_j} \right).
\end{IEEEeqnarray} 
The data-model~(\ref{eq_scch_2}) using (\ref{eq_log_membership3}) becomes as
 \begin{IEEEeqnarray}{rCl}
\nonumber \lefteqn{ \mu_{\mathrm{y}_j}(\tilde{\mathrm{y}}_j)} \\
\nonumber & \propto & \exp \left( - 0.5\beta \| \tilde{\mathrm{y}}_j\|^2  + \beta \left(\hat{m}_{\mathrm{u}_j }(\mathrm{y}_j)\right)^T (K_{\mathrm{a}\mathrm{a}})^{-1} (K_{\mathrm{x}\mathrm{a}})^T\tilde{\mathrm{y}}_j \right. \\
\nonumber && \left. {-}\: 0.5\beta \left(\hat{m}_{\mathrm{u}_j }(\mathrm{y}_j)\right)^T \left\{ (K_{\mathrm{a}\mathrm{a}})^{-1}  K_{\mathrm{x}\mathrm{a}}^T K_{\mathrm{x}\mathrm{a}}(K_{\mathrm{a}\mathrm{a}})^{-1}   + \frac{Tr(K_{\mathrm{x}\mathrm{x}}) - Tr((K_{\mathrm{a}\mathrm{a}})^{-1}  K_{\mathrm{x}\mathrm{a}}^T K_{\mathrm{x}\mathrm{a}})}{\nu + M - 2} (K_{\mathrm{a}\mathrm{a}})^{-1} \right\} \hat{m}_{\mathrm{u}_j }(\mathrm{y}_j) \right. \\
&& \left.  {-}\: 0.5\beta Tr\left( (K_{\mathrm{a}\mathrm{a}})^{-1} K_{\mathrm{x}\mathrm{a}}^T K_{\mathrm{x}\mathrm{a}} (K_{\mathrm{a}\mathrm{a}})^{-1} \hat{K}_{\mathrm{u}_j}   + \frac{Tr(K_{\mathrm{x}\mathrm{x}}) - Tr((K_{\mathrm{a}\mathrm{a}})^{-1}  K_{\mathrm{x}\mathrm{a}}^T K_{\mathrm{x}\mathrm{a}})}{\nu + M - 2} (K_{\mathrm{a}\mathrm{a}})^{-1} \hat{K}_{\mathrm{u}_j} \right) \right).
\end{IEEEeqnarray} 
Thus, (\ref{eq_satguru_8}) follows.

\section{Membership-Mapping Output Estimation}\label{appendix_738242.626259}
Using (\ref{eq_pf_1001_student_t}) and (\ref{eq_pf_1002}), we have 
 \begin{IEEEeqnarray}{rCl}
\label{eq_satguru_3}\left< (\mathrm{f}_j)_i\right>_{\mu_{\mathrm{f}_j;\mathrm{u}_j } } & = & (K_{\mathrm{x}\mathrm{a}} (K_{\mathrm{a}\mathrm{a}})^{-1}  \mathrm{u}_j )_i \\
\label{eq_satguru_4}& = & G(x^i) (K_{\mathrm{a}\mathrm{a}})^{-1}  \mathrm{u}_j.
 \end{IEEEeqnarray}
Thus,   
 \begin{IEEEeqnarray}{rCl}
\label{eq_satguru_5} & \widehat{ \mathcal{F}_j(x^{i}) } = &  G(x^i)(K_{\mathrm{a}\mathrm{a}})^{-1} \left< \mathrm{u}_j \right>_{\mu^*_{\mathrm{u}_j}}.
 \end{IEEEeqnarray}  
Using (\ref{eq_satguru_7_5}) in (\ref{eq_satguru_5}), we have
\begin{IEEEeqnarray}{rCl}
\label{eq_satguru_11}  \widehat{ \mathcal{F}_j(x^{i})}   &=&\beta  \left(G(x^i) \right) (K_{\mathrm{a}\mathrm{a}})^{-1} \hat{K}_{\mathrm{u}_j} (K_{\mathrm{a}\mathrm{a}})^{-1} (K_{\mathrm{x}\mathrm{a}})^T \mathrm{y}_j .
 \end{IEEEeqnarray} 
Substituting $\hat{K}_{\mathrm{u}_j}$ from (\ref{eq_hat_K_u_1000}) in (\ref{eq_satguru_11}), we get (\ref{eq_satguru_12}).

\bibliographystyle{unsrt}  
\bibliography{bibliography}

\end{document}